\documentclass[runningheads]{llncs}

\usepackage[mobile]{eccv}

\usepackage{eccvabbrv}

\usepackage{wrapfig}

\definecolor{babyblue}{rgb}{0.54, 0.81, 0.94}
\definecolor{citrine}{rgb}{0.89, 0.82, 0.04}
\definecolor{misocolor}{rgb}{0.16,0.27,0.86}
\definecolor{jbcolor}{rgb}{0.9,0.4,0.2}
\definecolor{bernacolor}{rgb}{0.9608,0.4863,0.00}
\definecolor{carlcolor}{rgb}{0.0,0.9863,0.30}
\definecolor{grey}{rgb}{0.3, 0.3, 0.3}

\definecolor{graphicbackground}{rgb}{0.96,0.96,0.8}
\definecolor{rouge1}{RGB}{226,0,38}  
\definecolor{orange1}{RGB}{243,154,38}  
\definecolor{jaune}{RGB}{254,205,27}  
\definecolor{blanc}{RGB}{255,255,255} 
\definecolor{rouge2}{RGB}{230,68,57}  
\definecolor{orange2}{RGB}{236,117,40}  
\definecolor{taupe}{RGB}{134,113,127} 
\definecolor{gris}{RGB}{91,94,111} 
\definecolor{bleu1}{RGB}{38,109,131} 
\definecolor{bleu2}{RGB}{28,50,114} 
\definecolor{vert1}{RGB}{133,146,66} 
\definecolor{vert3}{RGB}{20,200,66} 
\definecolor{vert2}{RGB}{157,193,7} 
\definecolor{darkyellow}{RGB}{233,165,0}  
\definecolor{lightgray}{rgb}{0.9,0.9,0.9}
\definecolor{darkgray}{rgb}{0.6,0.6,0.6}
\definecolor{babyblue}{rgb}{0.54, 0.81, 0.94}
\definecolor{citrine}{rgb}{0.89, 0.82, 0.04}
\definecolor{misogreen}{rgb}{0.25,0.6,0.0}
\definecolor{PalePurp}{rgb}{0.66,0.57,0.66}
\definecolor{todocolor}{rgb}{0.66,0.99,0.99}
\definecolor{pearOne}{HTML}{2C3E50}
\definecolor{pearTwo}{HTML}{A9CF54}
\definecolor{pearTwoT}{HTML}{C2895B}
\definecolor{pearThree}{HTML}{E74C3C}
\colorlet{titleTh}{pearOne}
\colorlet{bull}{pearTwo}
\definecolor{pearcomp}{HTML}{B97E29}
\definecolor{pearFour}{HTML}{588F27}
\definecolor{pearFith}{HTML}{ECF0F1}
\definecolor{pearDark}{HTML}{2980B9}
\definecolor{pearDarker}{HTML}{1D2DEC}

\newcommand{\simclr}{\texttt{SimCLR}\xspace}

\newcommand{\beit}{\texttt{BEiT}\xspace}
\newcommand{\mae}{\texttt{MAE}\xspace}
\newcommand{\nnclr}{\texttt{NNCLR}\xspace}
\newcommand{\algo}{\texttt{BAM}\xspace}
\newcommand{\byol}{\texttt{BYOL}\xspace}
\newcommand{\swav}{\texttt{SwAV}\xspace}
\newcommand{\relic}{\texttt{ReLIC}\xspace}

\newcommand{\dino}{\texttt{DINO}\xspace}

\newcommand{\deit}{\texttt{DEiT}\xspace}
\newcommand{\moco}{\texttt{MoCo}\xspace}

\newcommand{\mocooo}{\texttt{MoCoV3}\xspace}

\newcommand{\supIn}{\texttt{Sup-IN}\xspace}

\newcommand{\LBFGS}{\normalfont\texttt{LBFGS}\xspace}

\let\originalleft\left
\let\originalright\right
\renewcommand{\left}{\mathopen{}\mathclose\bgroup\originalleft}
\renewcommand{\right}{\aftergroup\egroup\originalright}

\DeclareMathOperator*{\argmin}{arg\,min}

\newcommand*{\eqdef}{\triangleq}

\renewcommand{\epsilon}{\varepsilon}
\renewcommand{\tilde}{\widetilde}

\newcommand{\nothere}[1]{}

\usepackage{xspace}

\usepackage{graphicx}
\usepackage{booktabs}
\usepackage{xcolor,colortbl}
\usepackage{multirow}
\usepackage{amssymb}
\usepackage{pifont}
\newcommand{\cmark}{\ding{51}}%
\newcommand{\xmark}{\ding{55}}%

\usepackage[accsupp]{axessibility}  

\usepackage[pagebackref,breaklinks,colorlinks,citecolor=eccvblue]{hyperref}

\usepackage{orcidlink}

\begin{document}

\title{Unsupervised Representation Learning by \\Balanced Self Attention Matching} 


\author{Daniel Shalam \and Simon Korman}

\authorrunning{D. Shalam \and S. Korman}

\institute{Department of Computer Science, University of Haifa, Israel}

\maketitle

\begin{abstract}
Many leading self-supervised methods for unsupervised representation learning, in particular those for embedding image features, are built on variants of the instance discrimination task, whose optimization is known to be prone to instabilities that can lead to feature collapse. Different techniques have been devised to circumvent this issue, including the use of negative pairs with different contrastive losses, the use of external memory banks, and breaking of symmetry by using separate encoding networks with possibly different structures.
Our method, termed \algo, rather than directly matching features of different views (augmentations) of input images, is based on matching their self-attention vectors, which are the distributions of similarities to the entire set of augmented images of a batch. We obtain rich representations and avoid feature collapse by minimizing a loss that matches these distributions to their globally balanced and entropy regularized version, which is obtained through a simple self-optimal-transport computation.
We ablate and verify our method through a wide set of experiments that show competitive performance with leading methods on both semi-supervised and transfer-learning benchmarks. Our implementation and pre-trained models are available at \url{github.com/DanielShalam/BAM} .
\end{abstract}.

\vspace{-0.8cm}
\section{Introduction}
\label{sec:intro}

\begin{figure}[t!]
\centering
\noindent\hspace{-10pt}
\addtolength{\tabcolsep}{0.04cm}    
\begin{tabular}{c c}
\includegraphics[width=6.3cm]{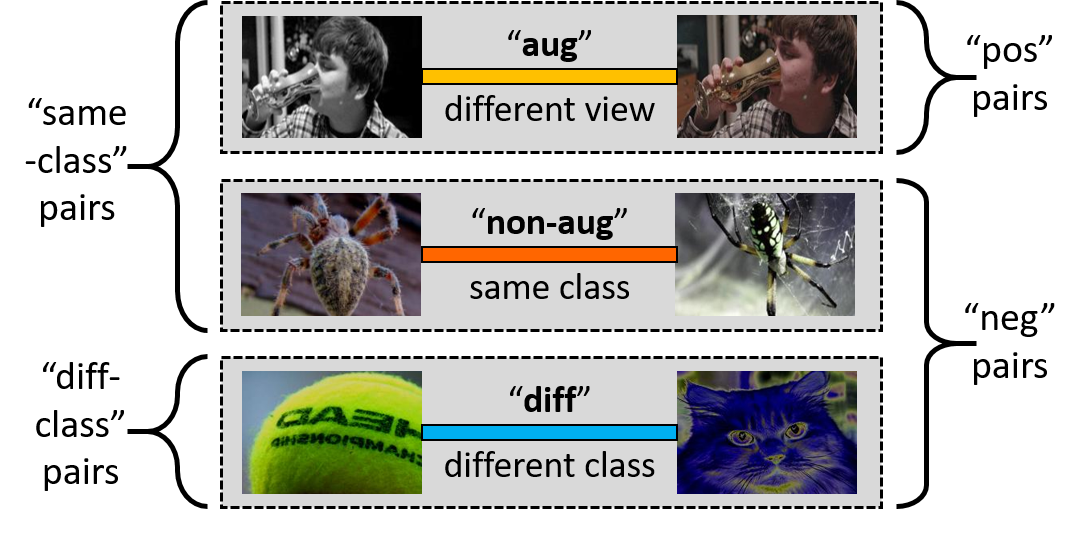} &  
\includegraphics[width=6.0cm]{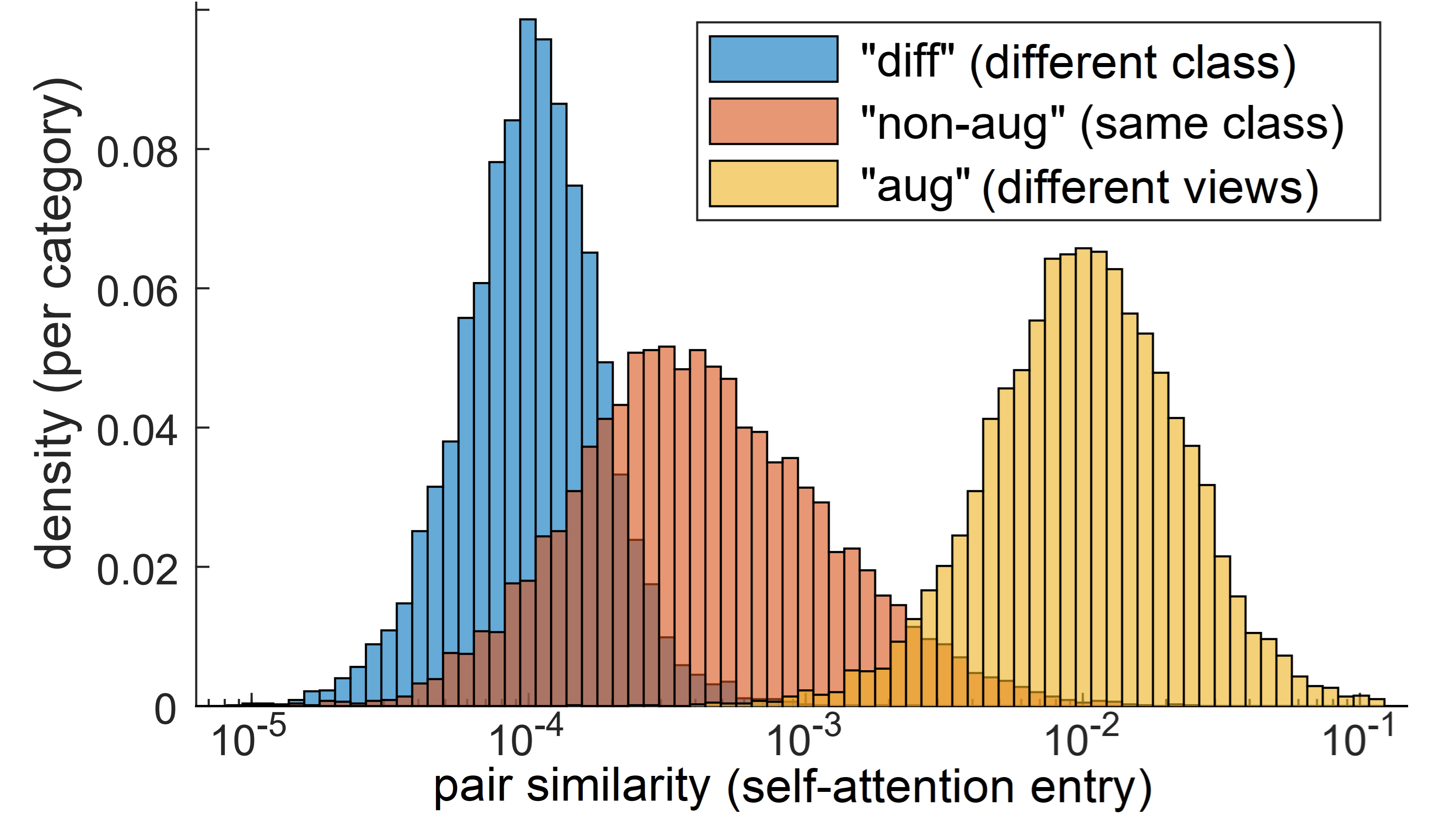}
\end{tabular}
\vspace{-10pt} 
\caption{
{\small \selectfont
{\textbf{Distributions of pairwise similarities in an augmented batch of images}. In 'instance-discrimination' self-supervised learning, pairs of instances (\textbf{left}) are typically categorized as "positives", which are pulled together in most approaches, or "negatives", which are pushed apart by contrastive approaches and totally ignored by others (momentum/distillation based). 
Our claim is that such simplistic same/not-same interpretations do not fully exploit the rich relative metric information that resides in the statistics of the entire set of pairwise similarities. For example, the categorization of pairs into "same-class" and "diff-class" (according to their class labels, which are unknown), shows (\textbf{right}) that while the similarities (soft-max entries) of "pos" pairs (yellow) are well separated from those of "diff" pairs (blue), the "non-aug" pairs similarities highly overlap the others. In our approach, the latents of a positive pair are pushed together by matching their self-attention distributions (which are the probabilities of the "neg" pairs they each belong to).
See Fig.~\ref{fig.paradigms} and text for further details.
}}}
\label{fig.pairs}  \vspace{-16pt} 
\end{figure}

Unsupervised representation learning has seen great success recently in a variety of machine learning applications that involve analysis and generation, especially in the domains of text and vision, and their combination. This is largely due to the emergence of large-scale self-supervised methods, which are designed to learn general-purpose feature extractors.

Their most attractive property is that the learned feature embeddings are not task specific, as they are trained on basic and unrelated pre-text tasks. Such representations have surpassed their supervised counterparts in their quality and speed of adaptation to previously unseen tasks, (i) by training a network head over the frozen feature extractor (a.k.a. \textit{probing}); (ii) by training the entire network using the well-initialized feature extractor (\textit{fine-tuning}); (iii) or by using only a small portion of data to do so, in a \textit{semi-supervised} setting.

Among the different lines of work on self-supervised visual representations, the most prominent paradigm is built on the idea of designing discriminative features that are invariant to distortion. Such methods train on a \textit{batch} of images, where random augmentations are used to produce several \textit{views} or \textit{instances} of each image. Different networks and loss functions are devised to solve a variety of instance discrimination pre-text tasks, that are based on relations between \textit{pairs} of instances in the augmented batch.

The self supervision typically amounts to a rather crude classification of the instance pairs into "positives", which are augmented views of the same image, and "negatives" - all the rest. See Fig.~\ref{fig.pairs} left, for an illustration showing one positive and two negative pairs of instances. While the different, perhaps more informative, classification of pairs as "same-class" or 'diff-class" is not available to the learner in the unsupervised setting, it clearly demonstrates the problem in treating the pos/neg classification in a binary same/not-same sense. 

\begin{figure}[t!]
\vspace{-3pt}
\hspace{-6pt}
\addtolength{\tabcolsep}{0.03cm}    
\begin{tabular}{c c c c}
\includegraphics[height=3.10cm]{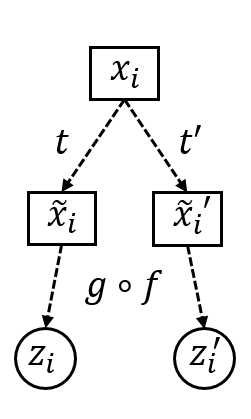} &  
\includegraphics[height=3.10cm]{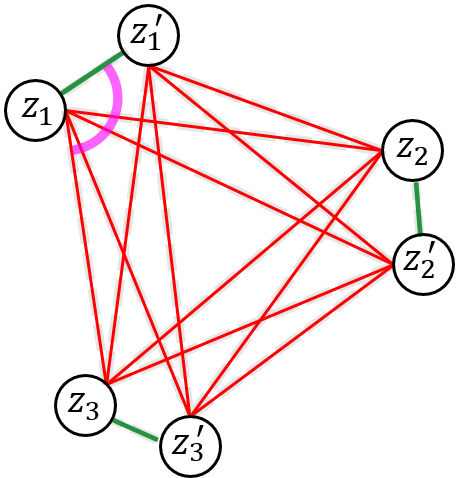} &  
\includegraphics[height=3.10cm]{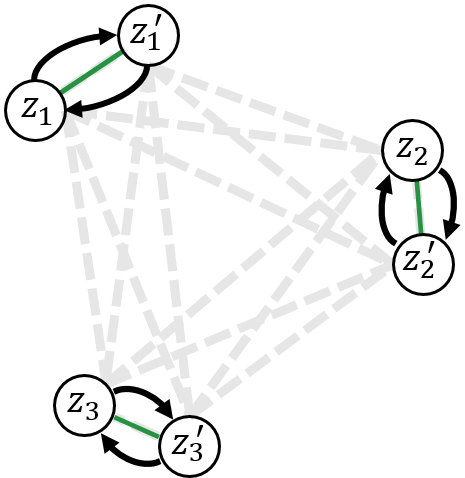} &  
\includegraphics[height=3.10cm]{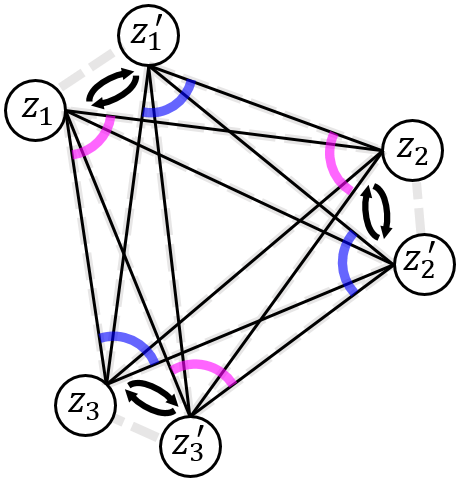}
\\ \vspace{5pt}
\textbf{(a)} img. to latents & \textbf{(b)} contrastive & \textbf{(c)} distillation & \textbf{(d)} self-attention \\
\end{tabular}
\vspace{-8pt} 
\caption{
{\small \selectfont
{\textbf{Paradigms of instance-discrimination based self-supervised learning} - demonstrated on a batch of 3 images $x_1,x_2,x_3$ with 2 augmentations. 
\textbf{(a)}: Each image $x_i$ is transformed into a pair of views by a pair of random augmentations. Both views are mapped to the latent space by a joint (learned) embedding. \textbf{(b)}: Contrastive-learning methods  compare each latent's attention distribution (marked by pink arc) to the 1-hot distribution of positive (green) and negative (red) matches.
\textbf{(c)}: Distillation methods focus on only positive pairs, by matching their latents under the online/offline encoders. 
\textbf{(d)}: Our approach suggests matching between the self-attention (SA) distributions of positive latent pairs. The SA distribution of one (pink arc) is matched to the balanced SA distribution of the other (blue arc). See text for detail.
}}}
\label{fig.paradigms}  \vspace{-14pt} 
\end{figure}

Figure~\ref{fig.pairs} right shows the distribution of actual pairwise similarity values, which are the entries of the large (8192$\times$8192) matrix of the pairwise cosine-similarities of latents of the entire set of augmented instances of a batch (of size 4096, with 2 augmentations) from the ImageNet~\cite{ILSVRC15} dataset (with soft-max per row that turns similarities to probabilities). The similarity values (probabilities) are shown separately for each of the 3 non-overlapping categories of pairs types: "diff-class", "non-aug" and "aug" (each histogram scaled to an area of 1, not in proportion to the category size). The feature embedding has relatively well separated the positive pairs from the different-class pairs, but the negatives distribution contains informative relative data, as shown by the sub-division into "non-aug" and "diff".

Instance-discrimination based approaches differ in how they operate on the graph of pairwise similarities between the latent representations of the instances, as can be seen in Fig.~\ref{fig.paradigms}, for an example batch of 3 images, with 2 augmentations. 

\textit{Contrastive} methods (like \simclr \cite{chen2020simple}) consider the entire similarities graph, pushing the positive and negative similarities to 1 and 0 respectively, or equivalently, pushing the yellow distribution in Fig~\ref{fig.pairs} to the far right and both blue and orange ones to the far left (zero). This approach suffers from the 'hard' same/not-same labelling, which often wrongly represents the more complex relations in the data.
\textit{Distillation} methods (like \byol \cite{grill2020bootstrap}) focus on matching the latents of only the positive pairs, using student-teacher type learning, and totally discard the subgraph of "negatives" that contains the vast majority of relative metric data contained in the batch.

We suggest a different approach of building on the characterization of positive pairs as ones that have similar \textit{self-attention} (SA) distributions of similarities to the entire augmented batch. It exploits information that was not used by other methods and leads to a simple method that avoids feature collapse and learns discriminative embeddings without the need for large memory banks, or queues of images, class prototypes, multiple network  structures, among others. 

Two important choices that we make can be seen in panel (d) of Figure~\ref{fig.paradigms}: (i) The self-attention (SA) distributions will include only "negative" pairs, resulting in more informative attention distributions that are not dominated by the "positives" (ii) The matching is between standard SA (i.e. soft-max) distributions (in pink) and globally-balanced entropy-regularized SA distributions (in blue), which guide the embedding to have self-attention relations with desired statistics and also avoid feature collapse.

In the experimental sections, we ablate over the suggested method and validate its empirical performance by evaluating it over standard benchmark protocols for linear probing, fine-tuning and semi-supervised learning on ImageNet~\cite{ILSVRC15} classification, transfer learning to classification tasks on other datasets, object-detection and instance-segmentation on COCO  \cite{coco} and semantic-segmentation on ADE20K \cite{ade20k}, as well as a very different application of video instance segmentation on DAVIS-2017 \cite{pont20172017}. 

Impressively, BAM strikes the best (and most stable) overall performance (comparing with ViT-B/16 architecture) over the two main evaluation protocols of ImageNet Classification (\textit{linear-probing} and \textit{fine-tuning}) with an average accuracy of $80.65$ compared to $80.5$ of \dino \cite{caron2021emerging} and $75.8$ of \mae \cite{he2022masked} (which are known to have inferior performance in fine-tuning and linear-probing respectively). \algo is also shown to be in par with \dino on semi-supervised and transfer learning image classification, and to surpass it on video instance segmentation.

\section{Related work}
\label{sec:related}

\subsection{Non-Discriminative Approaches}

This line of work focuses on learning meaningful representations without directly comparing different instances.
Inspired by the success of generative models in natural language processing (NLP), recent efforts have emerged aiming to learn representations by predicting masked patch tokens. 
For example, \beit \cite{bao2021beit} learns to predict discrete visual tokens of masked patches. Similarly, \mae \cite{he2022masked} employs a masked encoder-decoder ViT architecture  \cite{dosovitskiy2020vit} to predict pixel values, achieving state-of-the-art results across various recognition tasks. 
In contrast, the method proposed in this work learns feature representations by considering the intricate relations between the embeddings of a batch of samples, falling within the realm of discriminative approaches, which we delve into next.


\subsection{Discriminative Approaches}

Discriminative approaches in self-supervised learning employ methods that explicitly distinguish between features. For example, \cite{caron2018DeepCF, caron2020unsupervised, caron2021emerging, oquab2023dinov2} adopt a clustering-based strategy, discriminating features based on their clustering patterns. Specifically, \dino \cite{caron2021emerging,oquab2023dinov2} proposes a self-distillation approach wherein the base model learns to cluster akin to a teacher model. In subsequent work \cite{oquab2023dinov2}, this paradigm of self-distillation extended to larger models, resulting in robust linearly separated representations.


In a different direction, \simclr \cite{SimCLR} and \moco \cite{MOCO, MOCOv2, chen2021empirical} treat each image as a distinct class and train the model by discerning them through data augmentations. Typically, each image is treated as an individual class, and the model learns to classify each augmented image accordingly. These methods showed great success in learning meaningful representations. However, they also rely on large batches to generate ample negative samples.

Our approach shares similarities to this line of work as we operate on instances within a batch. However, we do not treat each image as a separate class and do not explicitly seek out negative samples.
For a comprehensive examination of the relationship between \algo and \simclr \cite{SimCLR} as well as \swav \cite{caron2020unsupervised}, please refer to the supplementary materials.


\section{Method}
\label{sec:method}



Our method operates, in each training iteration, on a batch of $n$ images $X=\{\mathbf{x}_1,\mathbf{x}_2,\dots,\mathbf{x}_n\}$. For each image $\mathbf{x}_i$ we synthetically generate $k$ random views by sampling augmentations $T_i=\{t_i^1,t_i^2,\dots,t_i^n\}$ from a distribution over image augmentations $\mathcal{T}$ (note that sampling is independent per image in the batch).
The result is a set of $n\times k$ augmented images, $\{t_i^j(x_i) : i\in[n]\:,\:j\in[k]\}$, for which we will use the shorter notation $\tilde{x}_i^j=t_i^j(x_i)$.

Following previous work, each augmented image $\tilde{x}$ first goes through a feature encoder $f$  , with resulting embedding $h=f(x)$, which is the target embedding that will be used in practice. Then, it is mapped by a projection head $g$ to a latent vector $z=g(h)$ which is used for loss computations during training. The networks $f=f_\theta$ and $g=g_\phi$ are parameterized by the parameter sets $\theta$ and $\phi$ over which optimization is jointly executed.
Let us denote the latent representation of the $j$th augmentation of the $i$th image by $z_i^j=g(f(\tilde{x}_i^j))$ and the entire set of latent embeddings of the augmented batch by 
    $Z=\{z_i^j : i\in[n]\:,\:j\in[k]\}$.

\subsection{Self-Attention Matching} \label{sec:SA_matching}

Contrastive methods learn the latent representation by pulling together 'positive' pairs of the form $z_i^{j}$ and $z_i^{j'}$ for $j\neq j'$ and pushing apart 'negative' pairs of the form $z_i^{j}$ and $z_{i'}^{j'}$ for $i\neq i'$.\footnote{Recall that subscripts refer to image index while superscripts refer to view index.} Likewise, distillation-based methods focus on the positive pairs only, matching the representations of student and teacher networks applied on the pair of different views of an image.

In contrast, our method learns by matching the \textit{self-attention distributions} of different views of an image. To do so, we calculate the following matrices: $S,A,B\in \mathbb{R}^{n\cdot k \times n\cdot k}$.
First, the matrix $S$, of all pairwise cosine similarities, that can be written as a $k\times k$ block matrix,
where each $n\times n$ block $S_{j,j'}$ is the pairwise cosine similarity matrix between views $j$ and $j'$. That is, $S_{j,j'}(i,i')=sim(z_{i}^{j},z_{i'}^{j'})$, where $sim(u,v)=u^T v/\|u\|\|v\|$ is the standard cosine similarity between vectors.

In order to represent each augmented image by its \textit{distribution} of similarities to the entire set of augmented images, we apply a soft-max operation with temperature parameter $\tau$ to each row of the matrix $S$, and obtain a right-stochastic\footnote{A matrix with non-negative entries and row sums of $1$ (representing distributions).} matrix $A$ (for 'attention'): \\ \vspace{-20pt}

\newcommand{\brows}[1]{%
  \begin{bmatrix}
  \begin{array}{@{\protect\rotvert\;}c@{\;\protect\rotvert}}
  #1
  \end{array}
  \end{bmatrix}
}
\newcommand{\rotvert}{\rotatebox[origin=c]{90}{$\vert$}}
\newcommand{\rowsvdots}{\multicolumn{1}{@{}c@{}}{\vdots}}

\vspace{-8pt}
\begin{equation}
\hspace{-20pt}
    S = \begin{bmatrix} 
    S_{11} & \dots  & S_{1k}\\
    \vdots & \ddots & \vdots\\
    S_{k1} & \dots  & S_{kk} 
    \end{bmatrix}
    \hspace{0.18cm} ; \hspace{0.18cm}   
    A = 
    \begin{bmatrix} 
    A_{11} & \dots  & A_{1k}\\
    \vdots & \ddots & \vdots\\
    A_{k1} & \dots  & A_{kk} 
    \end{bmatrix} = 
    \brows{a_1^1 \vspace{-4pt}\\ \rowsvdots\vspace{-3pt} \\ a_i^j\vspace{-3pt} \\ \rowsvdots\vspace{-3pt} \\ a_n^k}
    \hspace{0.18cm} ; \hspace{0.18cm}   
    a_i^j=\text{SM}_\tau(s_i^j)
    \label{eq.matrix_A}
\end{equation}

\noindent where $a_i^j$ and $s_i^j$ are the rows of $A$ and $S$ respectively that correspond to the $j$th view of the $i$th image and $\text{SM}_\tau$ is the soft-max\footnote{For any vector $v$ of length $M$, the $i$th coordinate of  $\text{SM}_\tau(v)$ is given by: $\exp(v_i/\tau)/\sum_{j=1}^M \exp(v_j/\tau)$. In this work we stick to the common setting of $\tau=0.1$.} operator with temperature $\tau$. 

Note that our attention computation is computed 'globally', once per batch, and not separately or 'locally' for each pair of augmentations (which would be a soft-max operation on each block $S_{j,j'}$), as is the case, \eg, in \relic \cite{mitrovic2020representation}.

Now that we have obtained the attention matrix that encodes for each latent in $Z$ its self-attention (SA) vector (distribution of similarities to the set $Z$), we proceed to design a loss that will enforce consistency of SA vectors across the different views (augmentations) of each image. This can be achieved by minimizing the distance between the respective distributions encoded by the rows of $A$, in a 'swapped' fashion, inspired by \swav \cite{caron2020unsupervised}, with the cross-entropy (CE) based (or equivalently, KL-divergence based) loss:
\begin{equation} \label{eq:loss_initial}
    \mathcal{L}_{\text{SA-matching}} =
    \sum_{j\neq j'}\sum_{i=1}^{n}\text{CE}(a_i^j,a_i^{j'})=
    -\sum_{j\neq j'}\sum_{i=1}^{n}\langle a_i^{j},\log a_i^{j'} \rangle
\end{equation}

As we demonstrate in our ablations, optimizing over this vanilla SA-matching loss function leads to undesirable trivial solutions that cause feature collapse. In the following, we suggest several additions and regularizations that will avoid feature collapse and further improve different qualities of the learned embedding.

\subsection{Focus on Negatives} \label{sec:focus_negatives}

Our SA matching loss \eqref{eq:loss_initial} is a sum of cross-entropy losses between pairs of corresponding rows of the self-attention matrix $A$. 
Following the naming conventions in Figure \ref{fig.pairs}, each row of $A$ contains a fraction of $1/n$ "pos" entries ($k$ out of $kn$), which are the similarities between a particular view and all ($k$) views of the same image. These entries reside on the diagonals of the block matrices of $A$, that are shown in \eqref{eq.matrix_A}. The remaining $(n-1)k$ row entries are the "neg"s, whose subdivision into "non-augs" (same class) and "diff" (different class) is unknown. 

Intuitively, since different views of a particular image are typically much more clustered together compared to arbitrary images of some class, the similarities associated with "augs" pairs are much higher than those of "non-augs" pairs. Indeed, as seen in the right panel of Figure \ref{fig.pairs}, the similarities (probabilities) of the "augs" pairs are very high, several orders of magnitude above the rest (notice the logarithmic $x$-scale). Therefore, they are very dominant in their contribution to the loss, at the expense of down-weighting the contributions of the "negative" matches, which provide the vast majority of attention information. Since we are matching between attention distributions, this fact introduces an undesired bias in the optimization in terms of the disproportionate importance of the sub-graph of 'augs' pairs, compared to the rest of the attention graph.
We therefore suggest to suppress the 'positives' altogether in the cross-entropy calculations. To do so, we 're-weight' the attention matrix $A$ by zeroing all 'augmentation' entries of the similarity matrix $S$, prior to the application of soft-max. Namely:

\vspace{-8pt}
\begin{equation}
    \hat{S}(i,j) = \begin{cases}
    0, & \text{i=j (mod n)}\\
    S(i,j), & \text{otherwise}
  \end{cases}
    \quad\quad ; \quad\quad
    \hat{a}_i^j=\text{SM}_\tau(\hat{s}_i^j)
    \label{eq.matrix_A_positives_only}
\end{equation}
where $\hat{a}_i^j$ and $\hat{s}_i^j$ are the rows of new augmentation-suppressed matrices $\hat{A}$ and $\hat{S}$  that correspond to the $j$th view of the $i$th image. For ease of notation, we will drop the $\;\hat{\cdot}\;$ (hat) notation in the following.

Note, importantly, that while the direct similarities between augmented views are not used anymore in the attention matching loss computation, the invariance to augmentation is enforced since we match the self-attention distributions, specifically, between augmented views. 

\subsection{Attention Balancing} \label{sec:balance}

We next suggest a regularization that breaks the symmetry in the loss \eqref{eq:loss_initial}\footnote{Symmetry, in the sense of comparing a set of distributions, the rows of $A$, to itself.}, by comparing the self-attention distributions, given by the rows of $A$, to a \textit{balanced} version of these distributions.
We introduce the 'target' matrix $B$ (for 'balanced'), which will be an optimal-transport regularized version of the soft-max (right-stochastic) matrix $A$. 
$B$ will be the closest\footnote{Closest in the sense of minimizing the KL-divergence $KL(B,A)$.} possible matrix to $A$, that is doubly-stochastic\footnote{A non-negative matrix with rows and cols that sum to 1 (representing distributions).} (which we refer to as 'balanced'). 

Given that $A$ is of the form $A=SM_\tau(S)$, it is a well known result that the matrix $B$ is the minimizer of the entropy-regularized functional:
    \begin{equation} \label{eq.OT_entropy_min}
        B=\argmin_{B'\in \mathcal{B}_{n\cdot k}}\:\langle -S,B'\rangle-\tau h(B')
    \end{equation} 
where $\mathcal{B}_{n\cdot k}$ is the set (known as the Birkhoff polytope) of $n\cdot k\times n\cdot k$ doubly-stochastic matrices, $\langle\cdot,\cdot\rangle$ stands for the Frobenius (standard) dot-product, $h(M)=-\sum_{i,j} m_{ij} \log(m_{ij})$ is the Shannon entropy of a matrix $M$ and $\tau$ (in our case the soft-max temperature) is the entropy regularization parameter. 

The interpretation of the OT formulation is that the matrix $B$ is the (optimal) transport-plan between a pair of length-$nk$ vectors of $1$s, with the negation of the similarity matrix $S$ acting as the transport cost matrix. Each row of $B$ 'transports' a total amount of 1 (that is the sum of similarities of the respective feature), by distributing it across the row entries, trying to minimize the transportation cost (given by the dot-product with the negation of the pairwise similarity matrix $S$). But importantly, this is not done independently per row (as happens with soft-max), but rather simultaneously for all rows, with the constraint that each column sums to one, meaning that each column 'receives' the same total of amount of 1, from all of the rows. 

In addition, The entropy regularization term provides means for controlling the entropy of the transport plan $B$. The smaller $\tau$ is, the lower the entropy is, resulting in sparse (quasi-deterministic) solutions that are 
difficult to optimize with respect to \cite{cuturi2013sinkhorn}. In contrast, larger values of $\tau$ give a large support that can represent rich pairwise feature relations.

The functional in \eqref{eq.OT_entropy_min} is strictly convex and  can be approximated very efficiently with matrix scaling algorithms, such as the Sinkhorn-Knopp method \cite{cuturi2013sinkhorn}, which typically requires very few linear-time iterations to converge. It amounts to alternating between row-normalization ($K=K./sum(K,\text{'rows'})$) and column-normalization ($K=K./sum(K,\text{'cols'})$) for an initialized $K=\exp(S/\tau)$. Notice that the soft-max matrix $A$ is the result of the first row normalization.
Another important property that we utilize, due to our particular case that $S$ consists of \textit{self} pairwise similarities (between the set of features $Z$ and itself), is that the symmetry of $S$ implies the symmetry of $B$ (see e.g. \cite{zass2006doubly}), while $A$ might be far from being symmetric. 

In summary, we rewrite the loss function by replacing the 'target' matrix $A$ in the objective of \eqref{eq:loss_initial} with the balanced and entropy-regularized matrix $B$:
\begin{equation} \label{eq:loss_final}
    \mathcal{L}_{\text{BAM}} = \sum_{j\neq j'}\sum_{i=1}^{n}\text{CE}(b_i^{j},a_i^{j'})=
    -\sum_{j\neq j'}\sum_{i=1}^{n}\langle b_i^{j},\log a_i^{j'} \rangle
\end{equation}


\subsection{Entropy Control} \label{sec:entropy}

In a self-supervised setting, entropy plays a crucial role in achieving a balance between exploration and exploitation during the learning process. The entropy of $A$ is typically controlled by dividing the pairwise similarities by
%
\begin{wrapfigure}{r}{6.3cm}
\vspace{-18pt}
\centering
\includegraphics[height=4.2cm]{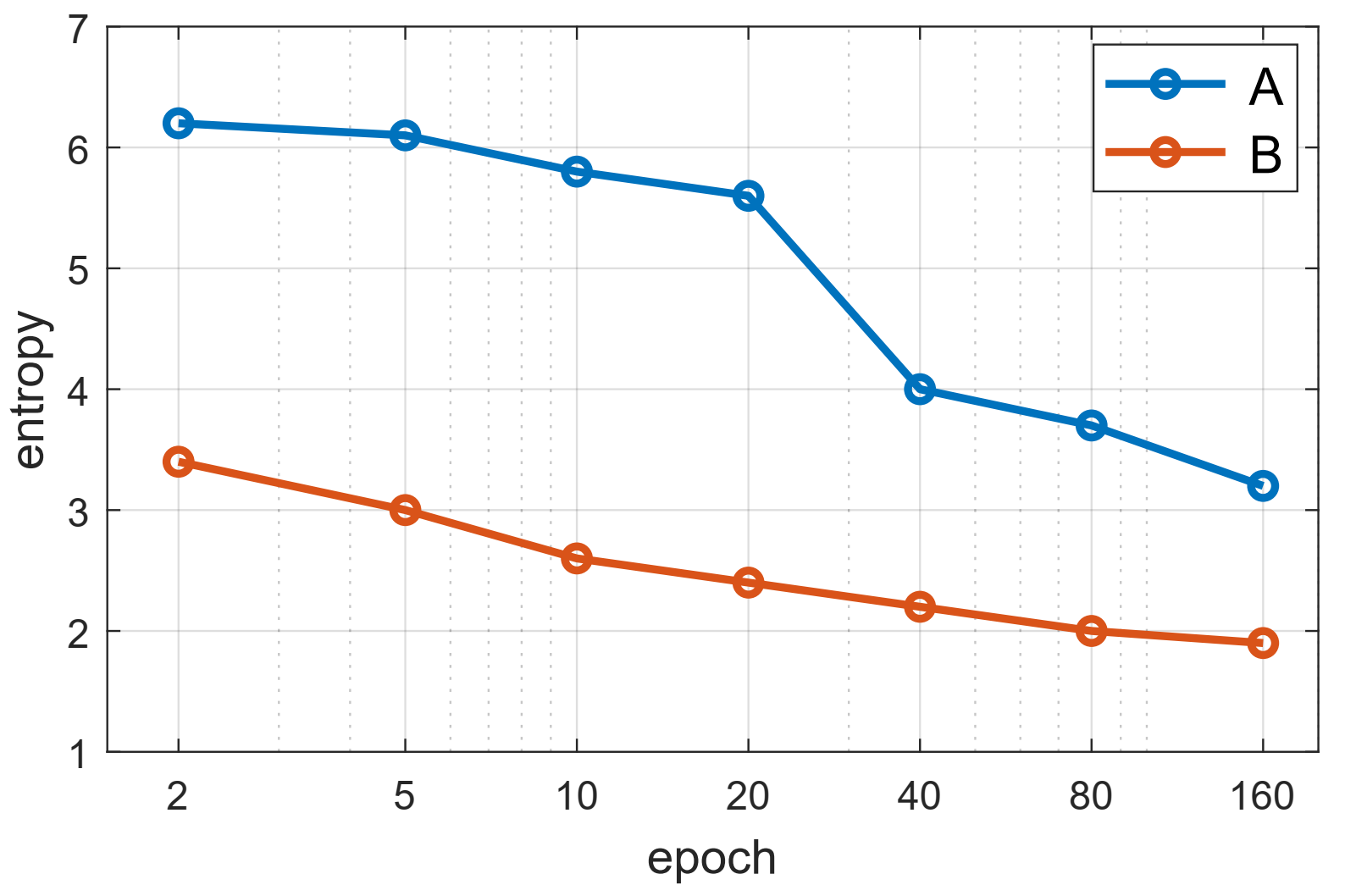}
\vspace{-22pt} 
\caption{
{\small 
{\textbf{Evolution of entropy} at training.
}}}
\label{fig.entropy_symmetry_progression}  \vspace{-20pt} 
\end{wrapfigure} 
%
the temperature parameter $\tau$, usually set to a value smaller than 1 (typically $0.1$), before applying the soft-max function. As we are learning without labels, excessive exploration (i.e., high entropy) may result in a uniform distribution among the samples in the batch. To address this, we use a temperature gap between $A$ and $B$, by using a separate target temperature parameter $\tau_B=0.05$, to be smaller than that of $A$ (half the standard level of $\tau=0.1$). This ensures that the target entropy always remains lower, attracting that of $A$ downwards, thereby facilitating exploitation as the training progresses.
We report in Fig.~\ref{fig.entropy_symmetry_progression} the progression of entropy of both matrices $A$ and $B$ during training. As we would have expected, we observe clearly how the entropy decreases consistently, while the gap keeps remaining and therefore attracting the entropy of $A$ towards that of $B$.


\subsubsection{Feature Collapse Avoidance}
There are two main forms of feature collapse, widely mentioned in the literature \cite{chen2020simple,grill2020bootstrap,caron2021emerging,zbontar2021barlow}, that are typical to instance-discrimination methods. The first, more common, happens when obtaining the trivial solution that maps the image space to approximately a constant latent vector. In the second, the embedding is similar to a random projection and the features are spread in space, approximately uniformly. Both cases include some form of view invariance and hence need to be avoided by additional mechanisms, such as the separation between negatives \cite{chen2020simple}, feature redundancy reduction \cite{zbontar2021barlow}, breaking the symmetry across augmented encodings in terms of structure (e.g. online predictor head, centering and sharpening \cite{caron2021emerging}) or momentum encoding \cite{grill2020bootstrap}.

In our method, both types of trivial solutions are avoided, as we intuitively explain next. Recall that our method pulls the self-attention matrix $A$ towards being balanced, \ie bi-stochastic (and symmetric). These properties directly contradict the situation where all features are closely concentrated in some small cluster: The attention matrix $A$ will be highly non-symmetric, since the similarity between a feature in the cluster center and a feature that is further out, will be weighted differently by the soft-max operations in the two respective rows, due to the higher sum of similarities of the central feature. This contradicts the bi-stochasticity and the symmetry that it guarantees. 

The other kind of trivial solution, where points are randomly spread out, might adhere to the balancing constraints, however, such solutions imply extremely uniform similarity and attention matrices, of very high entropy. Our mechanism of reducing the entropy of $A$ by maintaining a lower entropy in the target matrix $B$ leads the embedding to produce highly non-uniform attention maps, of relatively low entropy.


\subsection{Method Ablation}
In Table \ref{table:ablations} we ablate the individual contributions of several main ingredients of our method, by comparing linear probing accuracies on ImageNet~\cite{ILSVRC15}. 
The baseline \algo (top row) is our full configuration where a ViT-S/16 network is trained with batch-size of 4096 for 150 epochs. These settings are kept fixed for the other versions, which drop the following ingredients individually. 

\begin{itemize}
    \item \textit{\textbf{pos. masking}} refers to our decision of computing the self-attention based only on "negative" pairs (see Sec. \ref{sec:focus_negatives} and Eq.~\eqref{eq.matrix_A_positives_only} in particular). Removing this option amounts to cancelling the zero-masking of the relevant diagonals in the similarity matrix $S$ (line 14 of Alg. 1 in the supplementaries).
    \item \textit{\textbf{global norm}} refers to our method of constructing one single large ($nk\times nk$) matrix and performing normalizations (whether the stochastic soft-max, or bi-stochastic Sinkhorn) only \textit{once}, in a global manner (see Sec. \ref{sec:SA_matching} and Eq.~\eqref{eq.matrix_A} in particular). The standard alternative (\eg in \swav~\cite{caron2020unsupervised} and \relic~\cite{mitrovic2020representation}) would be to separately normalize each of the $k^2$ blocks (of size $n\times n$) of the matrices $A$ and $B$ (\ie for each pair of augmentations considered).
    \item \textit{\textbf{attn. balancing}} refers to the balanced normalization that we apply to the target matrix (see Sec. \ref{sec:balance}). Without this balanced target matrix $B$ (obtained from $S$ using Sinkhorn iterations), the loss function used contains only rows of the matrix $A$, as in Eq. \eqref{eq:loss_initial}.
    \item \textit{\textbf{multi-crop}} refers to the use of a mix of views with different resolutions in place of two full-resolution views \cite{caron2020unsupervised}. Our implementation uses the common setting of 2 large 'global' views and 6 small 'local' views, where the loss function is computed over all pairs of augmentations except 'local'-'local' pairs. Removing the multi-crop setting simply means using 2 large views, as given in the pseudo-code Alg. 1 in the supplementary materials.
    \item \textit{\textbf{stud-teacher}} 
    refers to the utilization of a teacher in the loss computation. 
    The parameters of the teacher network are updated using an exponential moving average (EMA) of the student network's parameters, ensuring that gradients propagate solely through the student. While our approach draws inspiration from techniques like \moco\cite{chen2021empirical}, it is tailored to suit the unique attributes of \algo. 
    In the supplementaries we provide further details.
    \item \textit{\textbf{contr. crop}} refers to the use of ContrastiveCrop \cite{peng2022crafting} - a semantic-aware localization of random crops that assists in avoiding both false positive and trivial positive pairs. We extend the technique of \cite{peng2022crafting}, that was proposed for CNN backbones, to work with Vision Transformers, basically, by computing region saliency based on ViT's informative attention maps rather than the convolutional feature maps. 
    
\end{itemize}

The results validate the separate contribution of each of the main parts of the method that were considered. Most notably, the use of attention balancing is shown to be critical for avoiding feature collapse, as it is the only mechanism that our method has (and needs) in this respect. In addition, the global normalization is shown to be a critical factor for the success of the proposed method.

\begin{table}[t]
\vspace{-2pt}
\hspace{-5pt}
\setlength\tabcolsep{2pt}
    \small
    \centering
    \fontsize{7.5}{7.5} \selectfont
    \renewcommand{\arraystretch}{1.18}
    \setlength{\tabcolsep}{0.17em} 
    \begin{tabular}{|c|c|c|c|c|c|c|}
    \hline
    {\textit{\textbf{pos. masking}}} & 
    {\textit{\textbf{global norm}}} & 
    {\textit{\textbf{attn. balance}}} & 
    {\textit{\textbf{multi-crop}}} & 
    {\textit{\textbf{stud-teacher}}} & 
    {\textit{\textbf{contr. crop}}} & 
    {\textit{\textbf{accuracy}}} \\ 
    \hline
    \cellcolor{pearDark!20}\cmark & \cellcolor{pearDark!20}\cmark & \cellcolor{pearDark!20}\cmark & \cellcolor{pearDark!20}\cmark & \cellcolor{pearDark!20}\cmark & \cellcolor{pearDark!20}\cmark & \cellcolor{pearDark!20}$\textbf{73.7}$ \\ 
    \hline
    \cellcolor{orange1!38}\xmark & \cmark & \cmark & \cmark &\cmark & \cmark & $72.6$ \\ 
    \hline
    %
    \cmark & \cellcolor{orange1!38}\xmark & \cmark & \cmark & \cmark & \cmark & $60.7$ \\ \hline    
    %
    \cmark & \cmark & \cellcolor{orange1!38}\xmark & \cmark & \cmark & \cmark & collapse \\\hline 
    %
    \cmark & \cmark & \cmark & \cellcolor{orange1!38}\xmark & \cmark & \cmark & $68.1$\\\hline
    %
    \cmark & \cmark & \cmark & \cmark & \cellcolor{orange1!38}\xmark & \cmark & $72.3$\\\hline
    %
    \cmark & \cmark & \cmark & \cmark & \cmark & \cellcolor{orange1!38}\xmark & $73.5$ \\\hline
    \end{tabular}
\vspace{6pt}
\caption{\small\label{table:ablations}\textbf{Ablations of method ingredients} using ImageNet~\cite{ILSVRC15} linear probing.
Models were trained for 150 epochs with ViT-S/16. 
}
\vspace{-22pt}
\end{table}

\vspace{-5pt}
\section{Main Results}
\label{sec:results}

\subsection{Implementation Details}

\newcommand{\augview}{augmented view}
\newcommand{\aaugview}{an augmented view}
\newcommand{\imageaugmentation}{image augmentation}
\newcommand{\Imageaugmentation}{Image augmentation}
\newcommand{\augmentation}{augmentation}
\newcommand{\todomga}[2][]{\todo[size=\scriptsize,color=red!20!white,#1]{MGA: #2}}
\newcommand{\dtheta}{{\delta\hspace{-.1em}\theta}}
\newcommand{\supervisedbaseline}{Supervised-IN}
\newcommand{\longalgo}{\textbf{B}ootstrap \textbf{Y}our \textbf{O}wn \textbf{L}atent}
\newcommand{\netparams}{{\theta}}
\newcommand{\targetparams}{\xi}
\renewcommand{\eqdef}{\mathrel{\ensurestackMath{\stackon[1pt]{=}{\scriptscriptstyle\Delta}}}}

\vspace{-0.4em}

A PyTorch-like pseudo-code of our method, for the case of $k=2$ augmentations, is provided in Algorithm 1 of the supplementary materials. Furthermore, we have made our code-base repository publicly available, to provide full reproducibility of the reported results. For lack of space, we report here results mainly for ViT architectures (Small and Base), while the respective tables with ResNet-50 results appear in the supplementaries.

\vspace{-6pt}
\paragraph{Training Setup.}
We perform self-supervised pretraining by training ResNet-50 \cite{ResNet} and ViT-Base \cite{dosovitskiy2020vit} models using the ImageNet dataset \cite{krizhevsky2012imagenet} without any labels. For ResNet-50, we employed the Lars optimizer \cite{LARS} with learning rate 0.3, which reached its peak value after 10 epochs. The training process utilized a batch size of 4096, distributed across 16 GPUs.
In the case of the ViT-Base model, we employed the AdamW \cite{ADAMW} optimizer with learning rate 0.003, which reached its peak value after 40 epochs \cite{Goyal2017}. The training process used a batch size of 4096. We followed the approach described in \dino \cite{caron2021emerging} and set the weight decay to follow a cosine schedule ranging from 0.04 to 0.4.
To ensure scalability and consistency, we adhere to the linear scaling rule \cite{Goyal2017}. After a warmup period, we gradually decay the learning rate using a cosine schedule \cite{SGDR}. 


\vspace{-6pt}
\paragraph{Optimization.}
The BAM loss \eqref{eq:loss_final} is minimized with respect to the parameters $\phi$ and $\theta$ of the embedding and projector networks $f$ and $g$. While both the source matrix $A$ and the target matrix $B$ depend on the these networks, we stop the gradients when computing $B$ from $A$ and use only its values, meaning that the gradients are back-propagated only through $A$'s side of the CE term in \eqref{eq:loss_final}. While the Sinkhorn procedure is fully differentiable and could be back-propagated through, we avoid doing so since we treat its output $B$ as target values that the self-attention matrix $A$ is trying to imitate. This decision also brings savings in computation. 

\vspace{-6pt}
\paragraph{Augmentation.}
Data augmentations, such as Gaussian blur, solarization, and color jittering, were employed following the techniques outlined in \byol \cite{guo2020bootstrap}. In addition, we incorporated the multi-crop strategy of \swav \cite{caron2020unsupervised} as well as the more recent idea of \textit{ContrastiveCrop} \cite{peng2022crafting}, that was designed for avoiding both false positive and trivial positive pairs of image crops.
While \textit{ContrastiveCrop} was originally proposed in the context of CNNs, we extend its applicability to Vision Transformer (ViT) backbones and find it advantageous, especially given our utilization of the Multi-Crop strategy, which inherently increases the likelihood of false positives due to the smaller size of individual crops.
%
%
We adapt \textit{ContrastiveCrop} to ViT by employing \textit{attention maps} from the final ViT block instead of utilizing \textit{feature maps}. Specifically, we compute attention scores between the [cls] token and all other patches, reshaping the resulting vector into an image-shaped heatmap. Since ViT's self-attention mechanism uses multiple heads, we treat each head as a feature channel, aggregating them by summation. Further details are in the supplementaries.
%

\subsection {Linear Evaluation on ImageNet}

We followed the standard evaluation protocol to assess the quality of \algo's representation. This involved training a linear classifier on top of the frozen representation, after removing the projector head.
We trained the linear classifier for 100 epochs using SGD and report top-1 accuracy in Table~\ref{table:linear_probing_IN}. 
For the ViT-S architecture, \algo achieved a top-1 accuracy of $75.0\%$ when trained for 300 epochs. This performance is competitive with previous self-supervised methods trained for the same duration and only second to \dino. Note that the \dino  ViT-S was also pre-trained for a longer period of 800 epochs (and used for the comparisons in the following experiments). We show both results here for a fair comparison. Regarding the ViT-Base architecture, \algo achieves comparable or better results compared
%
\begin{wraptable}{r}{6.8cm}
\vspace{-8pt}
    \small
    \centering
    \caption{\small\label{table:linear_probing_IN}\textbf{Linear Probing} on ImageNet}\vspace{-2pt}
     \renewcommand{\arraystretch}{1}
    \addtolength{\tabcolsep}{0.07cm}   
    \fontsize{8.9}{8.9} \selectfont
    \begin{tabular}[t]{l c c c c}
    \toprule
         \textit{\textbf{method}} & \textit{\textbf{arch.}} & \textit{\textbf{batch}} & \textit{\textbf{epochs}} & \textit{\textbf{acc.}} \\
         \midrule         
         \mocooo \cite{chen2021empirical} & ViT-S/16 & 4096 & 300 & ${73.2}$ \\
         \dino \cite{caron2021emerging} & ViT-S/16 & 1024 & 300 & $\bf{75.9}$ \\   
          \dino \cite{caron2021emerging} & ViT-S/16 & 1024 & 800 & $77.0$ \\
         \algo (ours) & ViT-S/16 & 4096 & 300 & $\underline{75.0}$ \\
         \midrule
         \mocooo \cite{chen2021empirical} & ViT-B/16 & 4096 & 300 & ${76.5}$ \\
         \dino \cite{caron2021emerging} & ViT-B/16 & 1024 & 400 & $\bf{78.2}$ \\
         \mae \cite{he2022masked} & ViT-B/16 & 4096 & 1600 & ${68.0}$\\        
         \algo (ours) & ViT-B/16 & 4096 & 300 & $\underline{78.1}$ \\
         \bottomrule                
\end{tabular}\vspace{-29pt}
\end{wraptable} 
to previous self-supervised approaches, with a top-1 accuracy of $78.1\%$. 
Moreover, when compared to methods that exclusively utilize in-batch relationships (e.g. \moco), \algo demonstrates a substantial improvement, indicating that its utilization of this information yields superior representations.


\vspace{-4pt}
\subsection {Fine-tuning on ImageNet}

While linear probing is commonly employed to evaluate self-supervised vision models, it primarily considers linear features and may overlook other important non-linear features. In our study, we opted to fine-tune our ViT models on ImageNet, utilizing the default hyperparameters outlined in \cite{deit} for a duration
\begin{wraptable}{r}{6.8cm}
\vspace{-28pt}
    \small
    \centering
    \caption{\small\label{table:IN_fine_tuning} \textbf{Fine tuning} on ImageNet.}\vspace{-4pt}
    \renewcommand{\arraystretch}{1}
    \fontsize{8.9}{8.9} \selectfont
    \addtolength{\tabcolsep}{0.08cm}  
    \begin{tabular}[t]{l c c c c}
    \toprule
         \textbf{\textit{method}} & \textbf{\textit{arch.}} & \textbf{\textit{batch}} & \textbf{\textit{epochs}} & \textbf{\textit{acc.}} \\
         \midrule
         %
         \mocooo \cite{chen2021empirical} & ViT-S/16 & 4096 & 300 & $\underline{81.4}$ \\
         \dino \cite{caron2021emerging} & ViT-S/16 & 1024 & 800 & $\bf{81.5}$ \\    
         \algo (ours) & ViT-S/16 & 4096 & 300 & ${81.3}$ \\
         \midrule
         \beit \cite{bao2021beit} & ViT-B/16 & 4096 & 300 & ${81.8}$ \\

         \dino \cite{caron2021emerging} & ViT-B/16 & 1024 & 300 & ${82.8}$ \\
         \mocooo \cite{chen2021empirical} & ViT-B/16 & 4096 & 300 & $\underline{83.2}$ \\
         \mae \cite{he2022masked} & ViT-B/16 & 4096 & 1600 & $\bf{83.6}$\\
         \algo (ours) & ViT-B/16 & 4096 & 300 & $\underline{83.2}$\\
         \bottomrule
    \end{tabular}
     \vspace{-19pt}
\end{wraptable} 
 of 150 epochs.
Table~\ref{table:IN_fine_tuning} reveals that our method, \algo, ranks second in performance, just behind \mae \cite{he2022masked}. Notably, \mae demonstrates the worst performance when evaluated using a linear classifier (Table \ref{table:linear_probing_IN}). Conversely, \algo outperforms \dino \cite{caron2021emerging} following the fine-tuning process. This observation suggests a contrasting trend compared to the linear (probing) evaluation.
By achieving strong performance in both linear evaluation and fine-tuning, \algo demonstrates its capability to generate robust non-linear features, that exhibit sufficient discriminative power to excel when employed with linear classifiers. 

\vspace{-4pt}
\subsection{Semi-supervised training on ImageNet} 
To further assess the performance of \algo, we conducted evaluations in the semi-
\begin{wraptable}{r}{7.6cm}
\vspace{-26pt}
    \fontsize{8.9}{8.9} \selectfont
    \centering
    \caption{\small\label{table:few_labels}\textbf{Semi-supervised training} on ImageNet.}\vspace{1pt}
    \setlength\tabcolsep{0.16cm}  
    \renewcommand{\arraystretch}{0.99}
    \begin{tabular}{l c c c c c c}
    \toprule
         \multirow{2}{*}{\textbf{\textit{method}}}
         & \multirow{2}{*}{\textbf{\textit{arch.}}} & \multirow{2}{*}{\textbf{\textit{params}}} & \multicolumn{2}{c}{Top-$1$} \\ 
                    &              &         & $1\%$ & $10\%$  \\ 
         \midrule
         
         
         
         
         
         
          
         \mocooo \cite{chen2021empirical}
         & ViT-S/16 & 
         $ 21   $M  & 
         $ 51.2 $   & 
         $ 58.7 $   & 
         \\
         \dino \cite{caron2021emerging}
         & ViT-S/16 & 
         $ 21   $M  & 
         $ \underline{58.7} $   & 
         $ \bf{73.9} $   & 
         \\
         \algo (ours)            
         & ViT-S/16 & 
         $ 21   $M  & 
         $ \bf{60.0}   $   & 
         $ \underline{73.2} $   & 
         \\
         \midrule
         \mocooo \cite{chen2021empirical}
         & ViT-B/16 & 
         $ 85   $M  & 
         $ \underline{66.3} $   & 
         $ 74.5 $   & 
         \\
         \dino \cite{caron2021emerging}
         & ViT-B/16   & 
         $ 85   $M    & 
         $ 65.0   $     & 
         $ \bf{76.0}   $     & 
         \\
         \mae \cite{he2022masked}
         & ViT-B/16 & 
         $ 85 $M    & 
         $ 57.4 $   & 
         $ 73.7 $   & 
         \\          
         \algo (ours)            
         & ViT-B/16 & 
         $ 85 $M    &
         $ \bf{66.8} $   &
         $ \underline{75.9} $   &
         \\
         \bottomrule
         
    \end{tabular}    
\label{table:semi_sup}
\vspace{-12pt}
\end{wraptable} 
%
supervised setup, by fine-tuning its representation on a classification task using a small subset of the labeled training data. The evaluation followed the semi-supervised protocol described in ~\cite{kornblith2019better,dmlab,SimCLR,CPCv2}, and utilized the same fixed splits of $1\%$ and $10\%$ of the labeled ImageNet training data as used in \cite{SimCLR}. 
We report top-1 accuracy in Table~\ref{table:semi_sup}, for the ViT-S and ViT-B architectures. 
We observed that \algo achieved a significant improvement ($+1.3\%$ on ViT-S) 
over other baselines when fine-tuned on ImageNet-$1\%$ using both architectures. In the case of 
over other baselines when fine-tuned on ImageNet-$1\%$ using both architectures. In the case of ImageNet-$10\%$, the performance of \algo remains competitive with other methods and is only surpassed by \dino (by a margin of $0.1\%$ on ViT-B).
These findings underscore the strong generalization capabilities of \algo, when applied to small subsets of the data.



\vspace{-5pt}
\subsection{Transfer to other classification tasks} 
\begin{table}[b!]
\vspace{-10pt}
     \caption{\small \textbf{Transfer learning classification by \textit{fine-tuning}} from ImageNet.}
     \label{tab:transfer_learning}
\centering
    \vspace{-9pt}
\renewcommand{\arraystretch}{0.99}
\setlength\tabcolsep{0.065cm}
    \fontsize{8.7}{8.7} \selectfont
\hspace{-0.1cm}
\begin{tabular}{l l c c c c c c c c}
\cmidrule[\heavyrulewidth]{1-10}
\textbf{\textit{protocol}} & \textbf{\textit{method}} & \textbf{\textit{arch.}} &
Food & CFR10 	 & 	CFR100 & 	Cars &  Aircraft &  Pets &  Flowers \\
\midrule
\multirow{4}{1.4cm}{\emph{Linear Eval.}}
& \simclr \cite{SimCLR}		 & 	ResNet-50 &
68.4 &	90.6	 &	71.6	 &	50.3 &	50.3	 &	83.6 &	91.2    \\
& \byol \cite{grill2020bootstrap}		 & 	ResNet-50 &
75.3 &	91.3	 &	78.4	 &	\textbf{67.8} &	60.6	 &	\underline{90.4} &	\underline{96.1}    \\
& \nnclr \cite{dwibedi2021little}		 & 	ResNet-50 &
\underline{76.7} &	\underline{93.7}	 &	\underline{79.0}	 &	\underline{67.1} &	\textbf{64.1}	 &	\textbf{91.8} &	95.1    \\
& \algo (ours)     & 	ResNet-50 &
\textbf{78.2} &	\textbf{94.3}	 &	\textbf{80.2}	 &	65.9 &	\underline{63.7}	 &	88.3 &	\textbf{96.3}    \\
\midrule
\multirow{4}{1.4cm}{\emph{Fine Tuning}} 
& \supIn \cite{dosovitskiy2020vit}  &	ViT-B/16 &
-- & 98.1		 & 	87.1
	 & 	-- &  --	 &  \underline{93.8} &  89.5    \\
& \moco-v3 \cite{chen2021empirical} &	ViT-B/16 &
-- & 98.9		 & 	90.5
	 & 	-- &  --	 &  93.2 &  97.7    \\
& \dino \cite{caron2021emerging} &	ViT-B/16 &
-- & \textbf{99.0}		 & 	\textbf{91.7}
	 & 	-- &  --	 &  -- &  \textbf{98.8}    \\
& \algo (ours)     & 	ViT-B/16 &
-- &	\textbf{99.0}	 &	\underline{91.4}	 &	-- &	--	 &	\textbf{93.9} &	\underline{97.9}  \\

\bottomrule
\end{tabular}

     \vspace{-4pt}
\end{table}

We further investigated the performance of \algo on transfer learning tasks involving other downstream classification tasks (over different datasets). For the ResNet-50 architecture, we conducted linear evaluation on a subset of classification tasks , following~\cite{SimCLR,kornblith2019better}. As for ViT, we followed the prevailing practice and fine-tuned our model end-to-end on the datasets used in \supIn \cite{dosovitskiy2020vit}. Accuracies in Table~\ref{tab:transfer_learning}, for both linear evaluation and fine-tuning, were obtained from on a held-out test set after hyperparameter selection on a separate validation set. 

During the linear evaluation, \algo exhibited significant improvements over other methods on four datasets, showcasing its superior performance. Notably, our approach achieved a $1.9\%$ improvement on the Food \cite{food101} dataset and a $1.5\%$ improvement on the CIFAR-100~\cite{cifar} dataset compared to the competing methods.
Regarding fine-tuning with ViT-B, our method demonstrated comparable performance to \dino. These findings indicate that the representations learned by \algo generalize effectively to unseen domains, further validating the effectiveness and adaptability of our approach.

\subsection{Video Instance Segmentation}

In Table \ref{table:video_seg}, we assess the performance of \algo on a dense recognition task. For the DAVIS-2017 video instance segmentation benchmark \cite{pont20172017}, we utilized \algo (frozen) output patch tokens, akin to the approach adopted in \dino \cite{caron2021emerging}. Following the protocol outlined in \cite{jabri2020space}, scenes were segmented by establishing nearest neighbor relationships between $T$ consecutive frames. We conducted experiments varying the value of $T$ within the range $[7, 14, 21, 28]$ and observed that as $T$ increased, \algo exhibited enhanced effectiveness compared to \dino (by $+0.9\%$). These results underscore the effectiveness of \algo in recognition tasks and its ability to leverage information from multiple frames more efficiently.

\begin{table}
\vspace{-12pt}
    \small
    \caption{\small\label{table:few_labels}
    \textbf{Video object segmentation} on DAVIS-2017. 
    We report mean region similarity \textit{J$_m$} and mean contour-based accuracy \textit{F$_m$}.
    Image resolution is 480p and $T$ is the number of frames used to establish nearest neighbor relations.
    }\vspace{-6pt}
    \setlength\tabcolsep{0.06cm} 
    \fontsize{6.8}{6.8} \selectfont
    \renewcommand{\arraystretch}{1.4}
    \begin{tabular}{l c | c c c| c c c| c c c| c c c}
    \toprule
\multirow{2}{*}{\textbf{\textit{method}}}
         & \multirow{2}{*}{\textbf{\textit{arch.}}} &          
        \multicolumn{3}{c|}{$T=7$}  &          
        \multicolumn{3}{c|}{$T=14$}  &          
        \multicolumn{3}{c|}{$T=21$} &          
        \multicolumn{3}{c}{$T=28$} \\
        & &
         {{$(\hspace{-0.04cm}J\hspace{-0.02cm}\&\hspace{-0.02cm}F\hspace{-0.02cm})\hspace{-0.02cm}_m$}}       & 
         {\textit{J$_m$}}     & 
         {\textit{F$_m$}}     &         {{$(\hspace{-0.04cm}J\hspace{-0.02cm}\&\hspace{-0.02cm}F\hspace{-0.02cm})\hspace{-0.02cm}_m$}}       & 
         {\textit{J$_m$}}     & 
         {\textit{F$_m$}}     &         {{$(\hspace{-0.04cm}J\hspace{-0.02cm}\&\hspace{-0.02cm}F\hspace{-0.02cm})\hspace{-0.02cm}_m$}}       & 
         {\textit{J$_m$}}     & 
         {\textit{F$_m$}}     &         {{$(\hspace{-0.04cm}J\hspace{-0.02cm}\&\hspace{-0.02cm}F\hspace{-0.02cm})\hspace{-0.02cm}_m$}}       & 
         {\textit{J$_m$}}     & 
         {\textit{F$_m$}}     \\
         \hline
         \dino \cite{caron2021emerging}
         & ViT-B/16 & 
         $ 62.3 $   &  $ 60.7 $   &  $ \bf{63.9} $ & 
         $ 62.8 $   &  $ 61.0 $   &  $ 64.7 $ & 
         $ 63.0 $   &  $ 61.2 $   &  $ 64.9 $ & 
         $ 63.1 $   &  $ 61.2 $   &  $ 65.0 $  
         \\  
         \algo 
         & ViT-B/16 &
         $ \bf{62.4} $   &  $ \bf{61.0} $   & $ \bf{63.9} $ &  
         $ \bf{63.3} $   &  $ \bf{61.7} $   & $ \bf{65.0} $ &  
         $ \bf{63.9} $   &  $ \bf{62.2} $   & $ \bf{65.5} $ &  
         $ \bf{64.0} $   &  $ \bf{62.3} $   & $ \bf{65.7} $   
         \\        
         \bottomrule         
    \end{tabular}    
\label{table:video_seg}
\vspace{-22pt}
\end{table}

\subsection{Object Detection, Semantic and Instance Segmentation}

Table \ref{table:seg_det} provides results of \textbf{object detection} and \textbf{instance segmentation} on the COCO dataset \cite{coco}, where the task layer is a Cascade Mask R-CNN \cite{cai2019cascade, he2017mask}, 
\begin{wraptable}{r}{7.5cm}
\vspace{-29pt}
    \centering
    \caption{\small\textbf{Object Detection, Semantic and Instance Segmentation} on COCO and ADE20K.}\vspace{-2pt}
    \setlength\tabcolsep{0.14cm}  
    \renewcommand{\arraystretch}{1.08}
    \addtolength{\tabcolsep}{-0.05cm}   
    \fontsize{8.7}{8.7} \selectfont
    \begin{tabular}[t]{l |c c |c c}
    \toprule
         \textit{\textbf{dataset}} & 
         \multicolumn{2}{c|}{\textit{\textbf{COCO}}} &
         \multicolumn{2}{c}{\textit{\textbf{ADE20K}}} \\
         \midrule
         \textbf{\textit{method}} &  
         \textbf{\textit{det. AP}} & \textbf{\textit{seg. AP}}  & 
         \textbf{\textit{m-IOU}} & \textit{\textbf{m-Acc.}} \\
         \midrule
         \mocooo  & ${47.9}$ & ${42.9}$ & ${-}$ & ${{-}}$ \\
         \beit          & ${49.8}$ & ${\underline{44.4}}$ & ${45.8}$ & ${{55.9}}$ \\   
         \mae         & ${\textbf{50.3}}$ & ${\textbf{44.9}}$ & ${-}$ & ${{-}}$ \\    
         \dino    & ${\underline{50.1}}$ & ${43.4}$ & ${\textbf{46.8}}$ & $\textbf{57.1}$ \\
         \algo 
         & ${\textbf{50.3}}$ & ${43.3}$ & ${\underline{46.1}}$ & $\underline{56.6}$ \\
         \bottomrule                
\end{tabular}\vspace{-6pt}
\label{table:seg_det}
\vspace{-12pt}
\end{wraptable} 
as was used in the baseline results. In addition, it includes results of \textbf{semantic} \textbf{segmentation} on the ADE20K dataset \cite{ade20k}, where we followed previous work
and used the UPerNet \cite{upernet} as the task layer.
These results add up to those of the additional \textbf{video object segmentation} experiment on Davis-2017, showing that the BAM has learned a representation that provides very competitive results on a wide range of tasks.

\vspace{-5pt}
\section{Conclusions}
\label{sec:conclusion}
\vspace{-6pt}
We introduced a novel instance-discrimination based approach for unsupervised representation learning. By leveraging a \textit{self-attention} matching mechanism, we demonstrated that a model can learn powerful and meaningful representations without explicitly treating samples as "positives" or "negatives". 
Our findings highlight the importance of considering the in-batch relationships that many approaches overlook, thereby capturing valuable information. 

The resulting method, termed BAM, was extensively validated on the standard SSL benchmarks, with highly competitive results for both linear-probing and fine-tuning (as opposed to methods that excel in one but fall short in the other) as well as in the semi-supervised, transfer-learning and object segmentation cases.

We believe our work will serve as a guide for future instance discrimination-based approaches, facilitating advancements in self-supervised learning.
\vspace{-4pt}

\bibliography{main}
\bibliographystyle{splncs04}


\definecolor{codegreen}{rgb}{0.0, 0.5, 0.0}
\newcommand\green[1]{\textcolor{codegreen}{#1}}
\newcommand\blue[1]{\textcolor{blue}{\textbf{#1}}}
\newcommand\red[1]{\textcolor{red}{\textbf{#1}}}

\appendix

%

\vspace{15pt}
\hspace{-22pt}
{\fontsize{16.1}{16.1} \selectfont {\textbf{Appendix}}}
\vspace{9pt}

{\fontsize{10}{10} \selectfont
\hspace{-18pt}
The Appendix includes the following sections:
\vspace{8pt}

\hspace{-3pt}\hyperref[sec:BAM_impl]{A. PyTorch-style BAM Implementation} 
\vspace{5pt}

\hspace{-3pt}\hyperref[sec:Arch_comparisons]{B. Comparison Across Architectures} 
\vspace{5pt}

\hspace{-3pt}\hyperref[sec:Arch_details]{C. Additional Architecture Details}
\vspace{5pt}

\hspace{-3pt}\hyperref[sec:Exp_details]{D. Additional Experiment Details}
\vspace{5pt}

\hspace{-3pt}\hyperref[sec:SA_vis]{E. Feature Self-Attention Visualizations}
\vspace{5pt}

\hspace{-3pt}\hyperref[sec:Entropy_reg]{F. A Note on Entropy Regularization}
\vspace{5pt}

\hspace{-3pt}\hyperref[sec:ViewSelect_vit]{G. Enhancing View Selection for ViT}
\vspace{5pt}

\hspace{-3pt}\hyperref[sec:Relation_prev]{H. Detailed Relations to Previous Work}
\vspace{5pt}

\hspace{-3pt}\hyperref[sec:Cluster_exp]{I. A Simple Clustering Experiment}
}

\section{PyTorch-style BAM Implementation}\label{sec:BAM_impl}

\begin{figure*}[h!]
\centering
\vspace{-2pt}
\includegraphics[width=0.99\textwidth]
{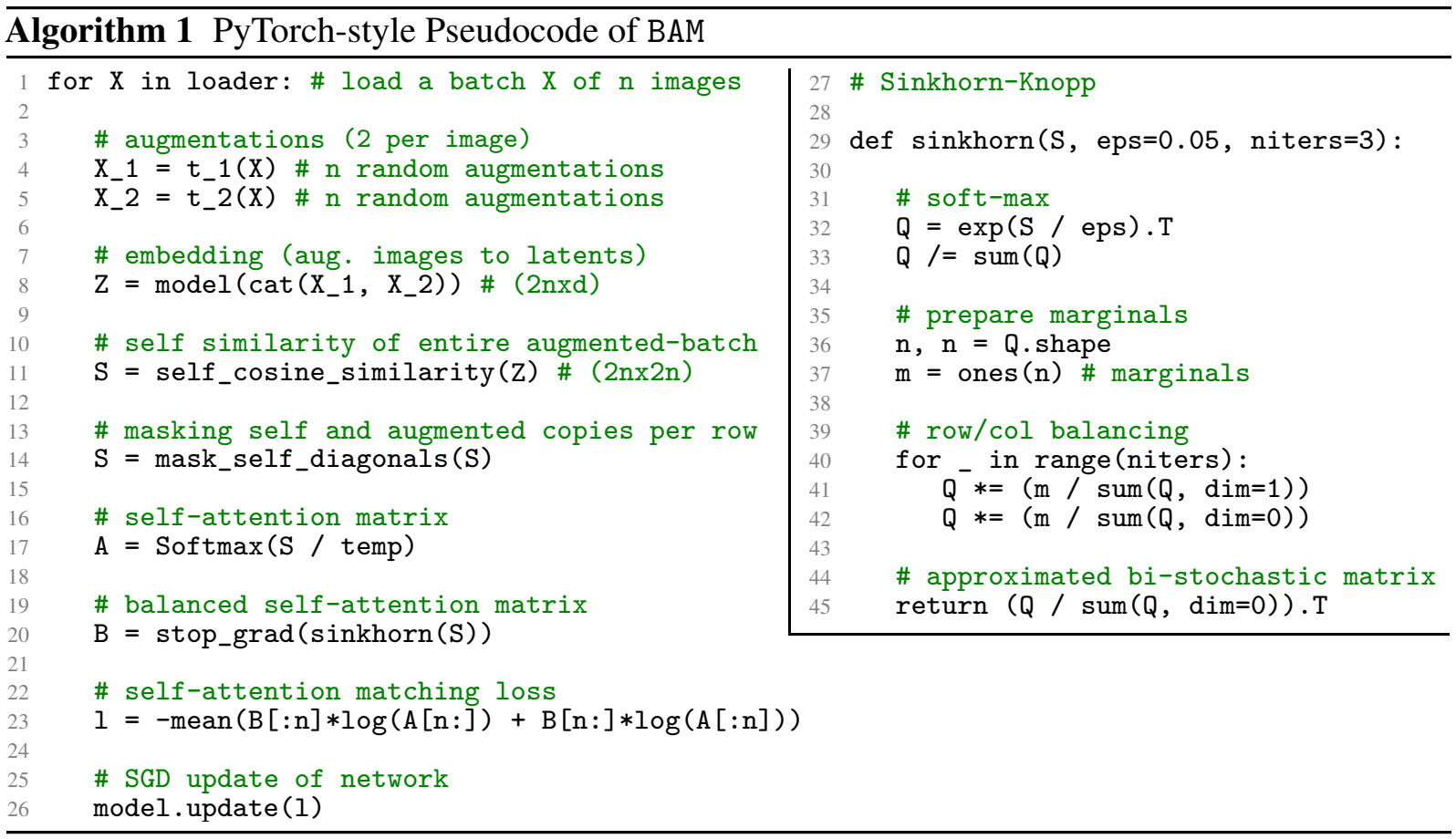}
\vspace{-6pt} 
\caption{
{\small \selectfont
{\textbf{Code snippet for model training given a batch of images}: The $k=2$ augmentations case.}}}
\label{fig.pseudo_code}  \vspace{2pt} 
\end{figure*}

Refer to \textbf{Algorithm 1} in Fig.~\ref{fig.pseudo_code} for a high-level implementation example of \algo.

\section{Comparison Across Architectures}\label{sec:Arch_comparisons}

In Table \ref{table:architectures}, we showcase the versatility of \algo across different backbone architectures, ensuring consistent computational requirements for fair comparisons. Our objective is to demonstrate the agnostic nature of \algo towards specific architectures. All backbones underwent rigorous training for 300 epochs on ImageNet-1k and were evaluated using linear evaluation protocols.
We chose the Resnet-50\cite{ResNet} architecture to compare against ViT-S and observed comparable results, with ViT-S showing a slight improvement (0.2\%). However, it's noteworthy that ResNet-50's performance could potentially be optimized further with a refined training strategy.
To evaluate against ViT-B, we selected the CAFormer-M36 architecture from \cite{yu2024metaformer}, incorporating both convolutional and attention blocks akin to ResNet and ViT. The same set of hyperparameters tuned to train ViT-B were used for CAFormer-M36 training. Our findings revealed a significant improvement (1.4\%) in linear probing accuracy while maintaining a lean parameter count.
This improvement underscores \algo's adaptability across diverse architectures, leveraging their unique characteristics to achieve performance gains.
It is important to emphasize that methods like \mae \cite{he2022masked} and \beit \cite{bao2021beit}, tailored specifically for ViT backbones, may face challenges when applied to different architectures, highlighting \algo's versatility and broader utility.

\begin{table}[h!]
\vspace{-4pt}
    \centering
    \caption{\small\label{table:few_labels}\textbf{Linear probing across architectures} on ImageNet.}\vspace{-1pt}
    \setlength\tabcolsep{0.12cm}  
    \renewcommand{\arraystretch}{0.99}
    \begin{tabular}{l c c c}
    \toprule & 
    \multirow{1}{*}{\textbf{\textit{arch.}}} &
    \multirow{1}{*}{\textbf{\textit{params}}} & 
    \multirow{1}{*}{\textbf{\textit{accuracy}}} \\
    \midrule

         & Resnet-50 & 
         $ 23 $M  & 
         $ 74.8\% $
         \vspace{2pt} \\ 
         
         & ViT-S/16 & 
         $ 21   $M  & 
         $ 75.0\% $
         \vspace{2pt} \\ 
         
    \midrule
    
         & ViT-B/16 & 
         $ 85 $M  & 
         $ 77.5\% $
         \vspace{2pt} \\ 
         
         & CAFormer-M36 & 
         $ 56   $M  & 
         $ \textbf{78.9\%} $
         \vspace{2pt} \\ 
         
         \bottomrule
         
    \end{tabular}    
\label{table:architectures}
\vspace{-6pt}
\end{table} 

\section{Additional Architecture Details}\label{sec:Arch_details}


\paragraph{\textit{Vision Transformer}.}
Our ViT architecture follows the implementation described in Dosovitskiy et al. \cite{dosovitskiy2020vit}. Specifically, we experiment with the ViT-B and ViT-S configurations, where the patch size is set to $16$. 
In addition to the patch embeddings, we include a generic [cls] token as an output feature, which is predicted along with the patch embeddings. 
To further enhance the stability of the training process, we adopt the sine-cosine variant of the positional embedding, suggested for self-supervised training in \cite{chen2021empirical}. 
%
To mitigate the high memory requirements and time complexity associated with the standard scaled-dot-product attention \cite{dosovitskiy2020vit}, we utilize a memory-efficient variant, as proposed by Rabe et al. \cite{rabe2021selfattention}. Furthermore, we leveraged mixed-precision training techniques to enhance both speed and memory utilization during training. Mixed-precision training allows us to perform computations using lower precision (such as FP$16$), which reduces memory requirements and speeds up computations. By employing those, we significantly reduces both the memory consumption and computational time, allowing us to achieve comparable performance to our ResNet-$50$ implementation. Specifically, We are able to run \algo with batch size of $4096$, while utilizing only two nodes, each equipped with 8 Tesla V$100$ GPUs. These optimizations enable us to balance computational efficiency and accuracy, facilitating the training of large ViT models within our limiting resource constraints.

\paragraph{\textit{Projector Head}.}
In our implementation, we employ a $3$-layer MLP for the projector heads across all configurations. The hidden layer and output layer dimensions are both set to 4096. 
Within the MLP, we utilize the rectified linear unit (ReLU) activation function \cite{nair2010rectified} for Resnet and GeLU\cite{hendrycks2016gaussian} for ViTs.
Specifically for ViT training, we observed that incorporating batch normalization after each layer significantly contributes to stabilizing the training process when we do not train with an EMA teacher.
When utilizing a teacher, we observed that no batch normalization is required, so the entire model is a \textit{BN}-free system.

\paragraph{\textit{EMA teacher}.}
While \algo demonstrates efficacy independently, integrating it  with a teacher network has proven to enhance stability and performance across downstream tasks. Several approaches can be adopted for this purpose. In our implementation, the teacher network computes the 'target' features.
Let's denote the set of latent representations produced by the model as $Z$ and the teacher representation as $\bf{Z_m}$. In the original implementation, we calculate our source similarity as $S=\text{Softmax}(\text{sim}(Z, Z))$ and the target matrix as $T=\text{Sinkhorn}(\text{sim}(Z, Z))$.
With the inclusion of the teacher network, the computation is modified to $S=\text{Softmax}(\text{sim}(Z, \bf{Z_m}))$ and $T=\text{Sinkhorn}(\text{sim}(\bf{Z_m}, \bf{Z_m}))$. 
This small change enables \algo to learn the feature distribution from \textbf{any} arbitrary teacher, enhancing its adaptability and performance.

\section{Additional Experiment Details}\label{sec:Exp_details}

\paragraph{\textit{Pre-Training} (done once, then used for all experiments)}
Hyperparameters for training \textbf{\algo} on the ImageNet dataset \cite{ILSVRC15} ("pre-training") with either Resnet or ViT are detailed in Table \ref{tab:configurations} (a).
\subsection{ImageNet Classification (paper Secs. 4.3-4.5)}

\paragraph{\textit{Linear Probing} (paper Sec. 4.3)}
Hyperparameters for training a \textbf{linear classifier} on frozen \algo features ("linear-classification") are presented in Table \ref{tab:configurations} (b).
For ViT, we use the concatenation of the [cls] token and average pooled patch tokens as the input to the classifier, similarly to \cite{caron2021emerging}.

\paragraph{\textit{Fine-Tuning} (paper Sec. 4.4)} 
We fine-tune our ViT models on ImageNet for $150$ epochs and use the  default hyper-parameters from the \deit \cite{deit} codebase. 

\paragraph{\textit{Semi-Supervised Fine-Tuning} (paper Sec. 4.5)}
Hyperparameters for fine-tuning the pre-trained model on a small labeled subset of ImageNet are presented in Table \ref{tab:configurations} (c).
In our semi-supervised experiments, we utilize the same data splits as those used in the work of \simclr \cite{chen2020simple}, specifically the same $1\%$ and $10\%$ labeled subsets of the ImageNet dataset. 
Using the ViT architecture, we adopt the SGD optimizer and train the model for $100$ epochs. We use the same default regularization techniques, such as mixup and cutmix, as provided in the codebase of DeiT \cite{deit}. We use a weight decay of $0.3$ and sweep the learning rate in each case. 
We use a single learning rate for both the main model and the linear head during fine-tuning. During training, we apply random resized crops to a size of $224$ and perform random horizontal flips on the training set. For validation, we resize the images to a size of $256$ and employ a center crop to a size of $224$.

\subsection {Compute requirements}

In Table \ref{table:compute}, we provide a breakdown of the time and memory requirements of \algo compared to other SSL methods. We implement them following their codebases. 
All methods were trained using the ViT-S/16 architecture across two nodes of A-100 GPUs (totaling 16 GPUs) for 100 epochs, with a batch size of 2048.
Our analysis reveals that the computational requirements of \algo are comparable to methods with similar architectural structures. Further details can be found in the supplementary materials.

\begin{table}[h]
\centering
\setlength\tabcolsep{4pt}
\renewcommand{\arraystretch}{1.1}
\vspace{-9pt}
\caption{\textbf{Experiment configurations}}
\vspace{2pt}
\begin{tabular}{|l|c|c|}
\hline
\textit{\textbf{config}} & 
\textit{\textbf{Resnet-50}} & 
\textit{\textbf{ViT-S/ViT-B}} \\
\hline
optimizer & LARS \cite{LARS} & AdamW \cite{ADAMW} \\ 
\hline
learning rate (LR) & 0.3 & 0.003 \\ 
\hline
LR schedule & cosine \cite{SGDR} & cosine \\
\hline
epochs & 300 & 300 \\ 
\hline
warmup epochs & 10 & 40 \\ 
\hline
weight decay & 1e-6 & 0.04-0.4 \\ 
\hline
batch size & 4096 & 4096 \\ 
\hline
gradient clipping & $-$ & 0.3 \\ 
\hline
large crops & 2 & 2 \\ 
\hline
small crops & 6 & 6 \\ 
\hline
\end{tabular}
\vspace{5pt} \\ 
{\small  \small{(a) ImageNet~\cite{ILSVRC15} pre-training}}\vspace{8pt}
\\

\begin{tabular}{|l|c|c|}
\hline
\textit{\textbf{config}} & 
\textit{\textbf{Resnet-50}} & \textit{\textbf{ViT-S/ViT-B}} \\
\hline
optimizer & SGD & SGD \\ 
\hline
learning rate & 0.3 & 0.002 \\ 
\hline
batch size & 1024 & 1024 \\ 
\hline
epochs & 100 & 100 \\ 
\hline
\end{tabular}
\vspace{5pt} \\ 
\small{(b) {ImageNet~\cite{ILSVRC15} Linear probing}}
\\ \vspace{8pt} 
\begin{tabular}{|l|c|}
\hline
\textit{\textbf{config}} & 
\textit{\textbf{ViT-S/ViT-B}} \\
\hline
optimizer & AdamW \\ 
\hline
learning rate & 5e-5 \\ 
\hline
batch size & 512 \\ 
\hline
epochs & 100 \\ 
\hline
warmup epochs & 5 \\ 
\hline
random erasing & 0.25 \\ 
\hline
drop path & 0.1 \\ 
\hline
mixup & 0.8 \\ 
\hline
cutmix & 1 \\ 
\hline
\end{tabular}

\vspace{3pt} 
\small{(c) {semi-supervised training on ImageNet~\cite{ILSVRC15} - $10\%$}}
\vspace{-14pt} 

\label{tab:configurations}
\end{table}
\begin{table}[b]
    \centering
     \caption{\small\label{table:compute}\textbf{Compute requirements}}\vspace{-2pt}
     \renewcommand{\arraystretch}{1}
    \addtolength{\tabcolsep}{0.03cm}   
    \fontsize{8.6}{8.6} \selectfont
    \begin{tabular}[t]{l c c c}
    \toprule
         \textit{\textbf{method}}    & \textit{\textbf{crops}}     & \textit{\textbf{time}}      & \textit{\textbf{mem./gpu}} \\ 
         \midrule  

         \moco\cite{chen2021empirical} &
         $2 \times 224^2$ &
         $5.5$h &
         $18.1$G \\

         \dino\cite{caron2021emerging} &
         $2 \times 224^2$ &
         $5.8$h &
         $18.6$G \\
         
         \algo (ours) &
         $2 \times 224^2$ &
         $5.4$h &
         $18.3$G \\
         
         \bottomrule                
\end{tabular}
\end{table} 

\begin{figure*}[t]
\hspace{-7pt}
\setlength\tabcolsep{0.05cm}  
\renewcommand{\arraystretch}{1.3}
\begin{tabular}{c c c}
\includegraphics[width=2.367785925cm] {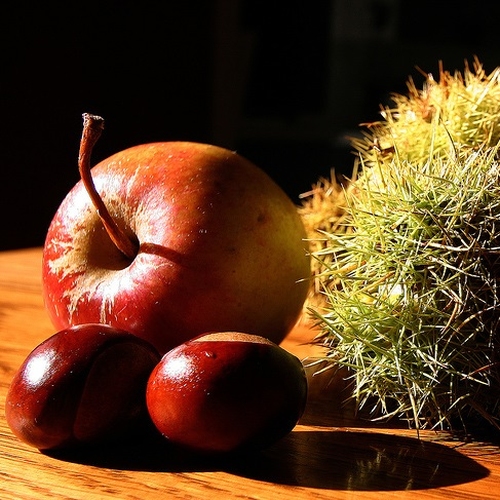}   &  
\includegraphics[width=2.367785925cm] {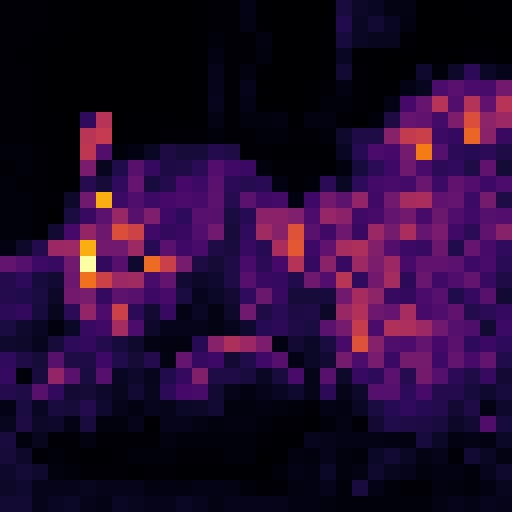} & 
\includegraphics[width=7.232004cm] {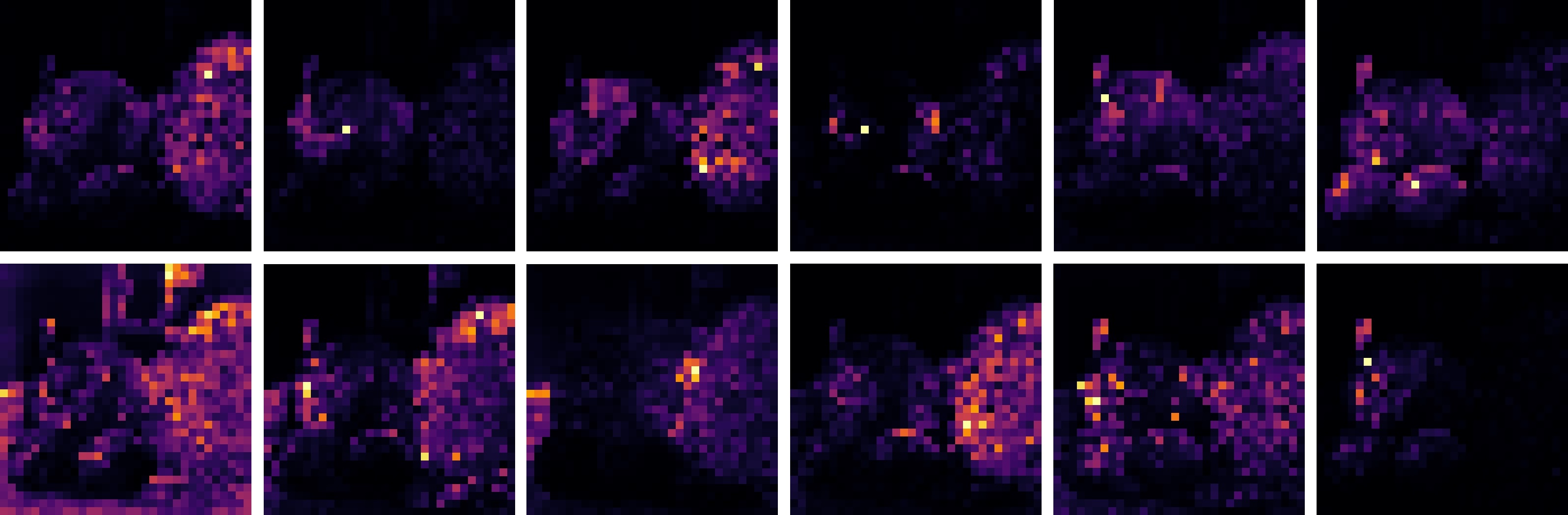}   \\   
\includegraphics[width=2.367785925cm] {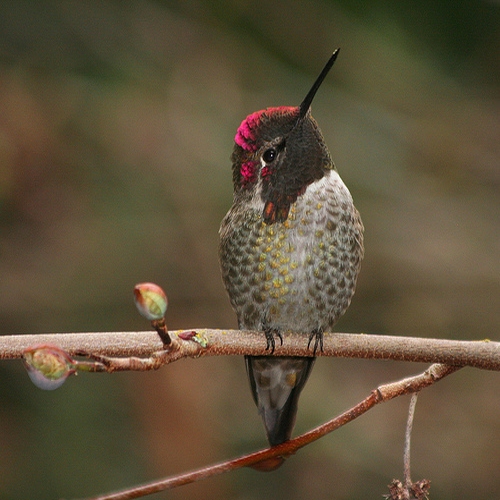}   &  
\includegraphics[width=2.367785925cm] {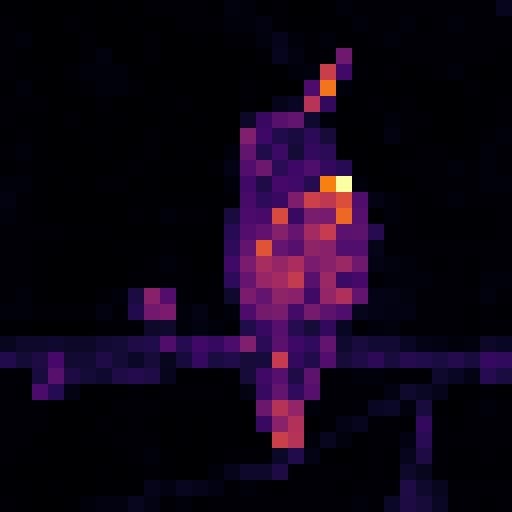} & 
\includegraphics[width=7.232004cm] {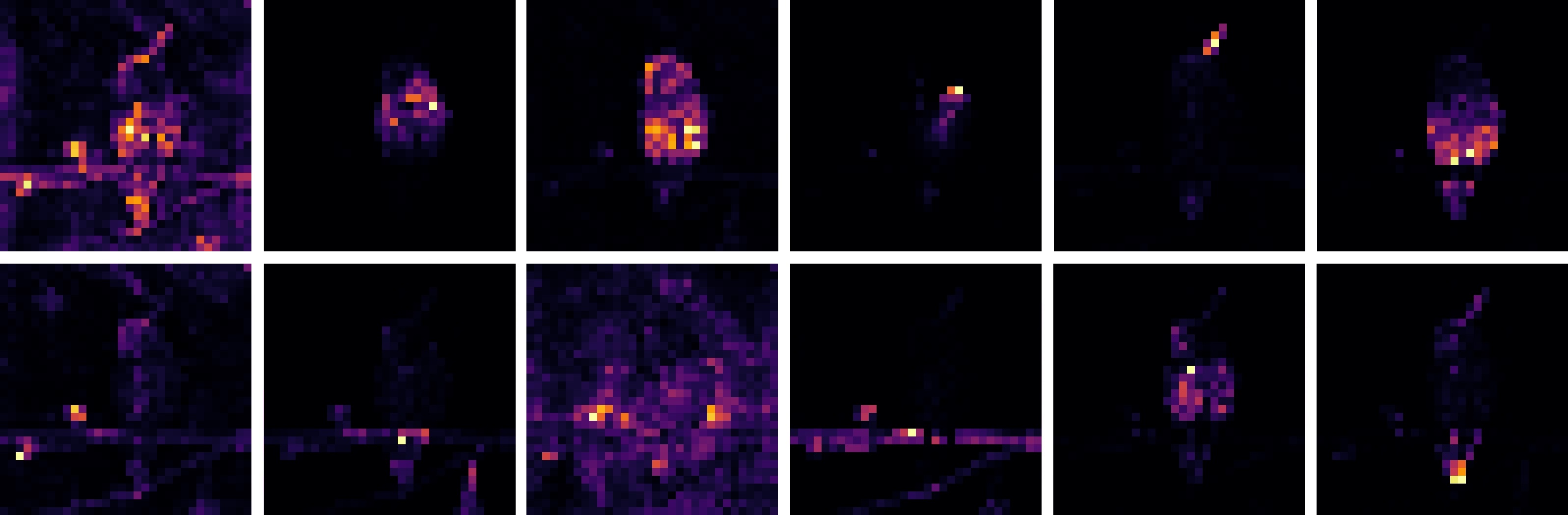}   \\   
\includegraphics[width=2.367785925cm] {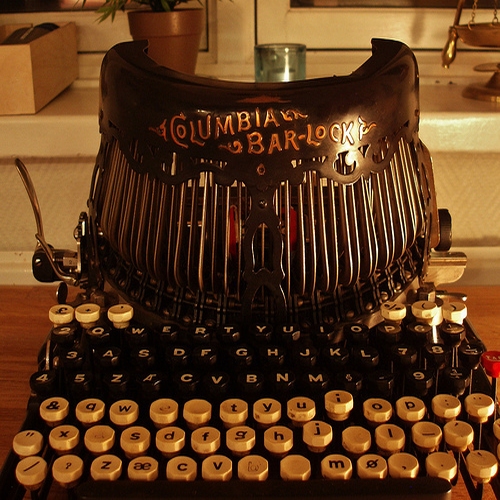}   &  
\includegraphics[width=2.367785925cm] {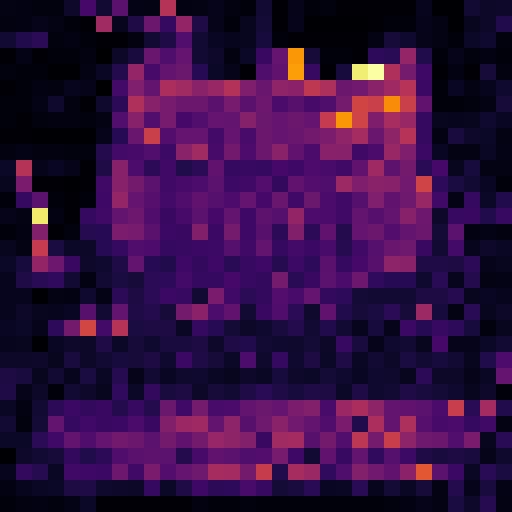} & 
\includegraphics[width=7.232004cm] {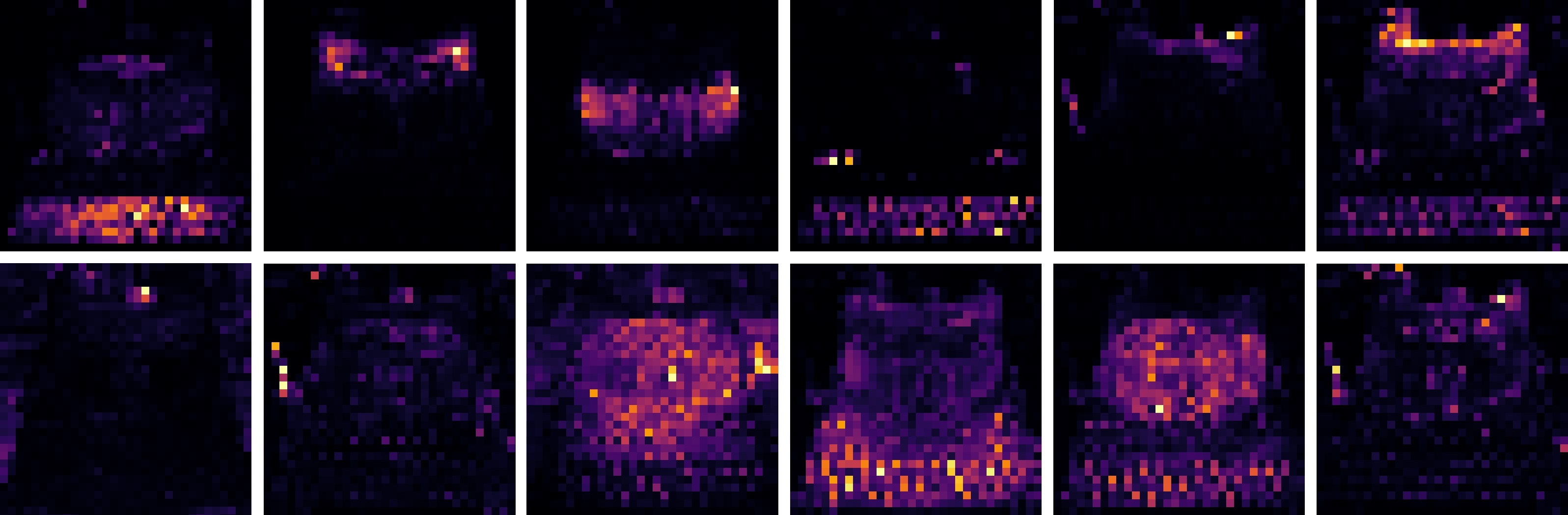}   \\  
\end{tabular}
\vspace{-9pt} 
\caption{
{\small \selectfont
{\textbf{Self-attention visualizations from the last layer of a \algo ViT-B network, that was pre-trained on the unlabeled ImageNet train-set. The three example images are from the validation set. To the right of each image is the attention map, which is the average of the attention maps produced by each of the 12 [cls] token heads, shown on the right.} }}}
\label{fig.SA_vis}  \vspace{-7pt} 
\end{figure*}

\subsection{Transfer Learning (paper Sec. 4.6)}
\paragraph{\textit{Transfer via Linear Classification} (Tab. 6 top in paper)}
To evaluate the transferability of our learned representations, we closely followed the linear evaluation protocol commonly employed in the literature \cite{kolesnikov2019revisiting, kornblith2019better, chen2021empirical, grill2020bootstrap}. Firstly, we freeze the pretrained parameters of our model and train a multinomial logistic regression classifier on top of the frozen representation. Importantly, during training and testing, we do not apply any image augmentations. Instead, we resize the images to a resolution of $224$ pixels along the shorter side using bicubic resampling and normalize them using ImageNet statistics. During the training of the linear classifier, we employ the \LBFGS optimization algorithm and minimize the cross-entropy objective with $\ell_2$-regularization. To determine the optimal regularization parameters, we explore a range of $45$ logarithmically-spaced values between $10^{-6}$ and $10^5$. We select the best-performing hyperparameters based on their performance on the validation set. Once the best parameters are chosen, we retrain the model on the combined training and validation images using the selected hyperparameters. The final accuracy is then reported on the test set.

\paragraph{\textit{Transfer via Fine-Tuning} (Tab. 6 bottom in paper)}
We investigate our ViT model transferability to other datasets by employing a fine-tuning approach. We adopt a fine-tuning protocol similar to that used in previous works such as \cite{chen2021empirical, dosovitskiy2020vit}, ensuring consistency in evaluation. To conduct the fine-tuning experiments, we utilize the same datasets and regularization strategies as mentioned in the aforementioned works. However, we make a modification to the learning rates originally used in \moco-v3 \cite{chen2021empirical} by multiplying them with a factor of $0.1$. This adjustment is motivated by our observation that our ViT-B/16 model tends to learn better when transferred to other datasets with smaller learning rates.

\section{Feature Self-Attention Visualizations}\label{sec:SA_vis}

In Figure \ref{fig.SA_vis} we provide self-attention visualizations for 3 different examples from the ImageNet-1k validation set.
The model is ViT-B, with images resized to $512\times512$ and patch size of 16.
For each image, we show the attention map to its right, which is the average attention map produced by each of the 12 [cls] token heads, shown on the right. 
It is apparent that the \algo attention maps automatically segment the input image in a fully unsupervised way, where each head is responsible for different patterns.

\section{A Note on Entropy Regularization}\label{sec:Entropy_reg}
In our training process, we made use of entropy regularization that is controlled by the $\lambda$ parameter of the Sinkhorn algorithm. Importantly, we set $\lambda$ to be less than the temperature parameter $\tau$ of the soft-max matrix, in order to encourage lower entropy of the soft-max attention matrix, as was explained in Section 3 of the paper and demonstrated in Figure 3 of the  paper.

We empirically explored this behaviour during our experiments, by training \algo with several values of $\lambda$, while fixing the commonly used value of $\tau=0.1$ \cite{chen2020simple}. We found that increasing $\lambda$ towards $\tau$ can lead to collapsed training, where the model fails to learn meaningful representations. Interestingly, the found "collapse threshold" (i.e. the maximum possible value for $\lambda$) is different for the two architectures. We observed collapsed training using $\lambda=0.075$ with Resnet and $\lambda=0.1$ with ViT. It is worth noting that our findings align with the results reported in \dino \cite{caron2021emerging}.


\section{Enhancing View Selection for ViT}\label{sec:ViewSelect_vit}

The selection of appropriate image views is paramount in capturing meaningful relationships between different parts of a scene or object. Traditionally, random cropping has been a prevalent technique for this purpose. However, recent work by \cite{peng2022crafting} has revealed its limitations, particularly in generating false positives, such as mistaking background elements for objects. In response, they introduced \textit{ContrastiveCrop}, a novel cropping method specifically designed to generate more accurate positive views for contrastive learning, particularly when coupled with \underline{ResNet} backbones.

While \textit{ContrastiveCrop} was originally proposed in the context of CNNs, we explore its applicability to Vision Transformer (ViT) backbones. We hypothesize that \textit{ContrastiveCrop} could be particularly advantageous for \underline{ViT} models, especially given our utilization of the Multi-Crop strategy, which inherently increases the likelihood of false positives due to the smaller size of individual crops.

The fundamental concept behind \textit{ContrastiveCrop} is to generate bounding boxes for each image in an unsupervised manner, ensuring that important details are encompassed within these regions. This is achieved by treating the output \textit{feature maps} of the network as heatmaps, which quantify the significance of individual pixels. These heatmaps are generated by aggregating (summing) features from the last convolutional layer across the channel dimension and normalizing the resulting sum to the range $[0, 1]$.

Adapting \textit{ContrastiveCrop} to ViT involves a straightforward modification. Instead of utilizing feature maps, we employ \textit{attention maps} from the final ViT block. 
Specifically, we compute attention scores between the [cls] token and all other patches, reshaping the resulting vector into an image-shaped heatmap. Since ViT's self-attention mechanism uses multiple heads, we treat each head as a feature channel, aggregating them by summing their values. 
Subsequently, we follow the same procedure outlined in the original paper \cite{peng2022crafting}.
It is worth noting that while the output of the attention maps is already a scaled heatmap of probabilities, alternative approaches such as using the cumulative distribution of pixel values might be more valid for ViT; however, we did not explore such options in this study.

During training, we initially employ \textit{RandomCrop} as the view selection strategy. After a warmup period of 40 epochs, we transition to \textit{ContrastiveCrop} and update bounding boxes every 20 epochs thereafter. It's noteworthy that we adhere to the hyperparameters prescribed in the original ContrastiveCrop paper without further modification.
We have observed promising results upon integrating our adaptation of \textit{ContrastiveCrop} into ViTs, particularly in conjunction with the utilization of smaller crops. This augmentation has demonstrated considerable potential, manifesting in enhanced training stability and improved performance across various downstream tasks.

\section{Detailed Relations to Previous Work}\label{sec:Relation_prev}
\begin{figure*}[t!]
\centering
\vspace{-5pt}
\includegraphics[width=0.99\textwidth]
{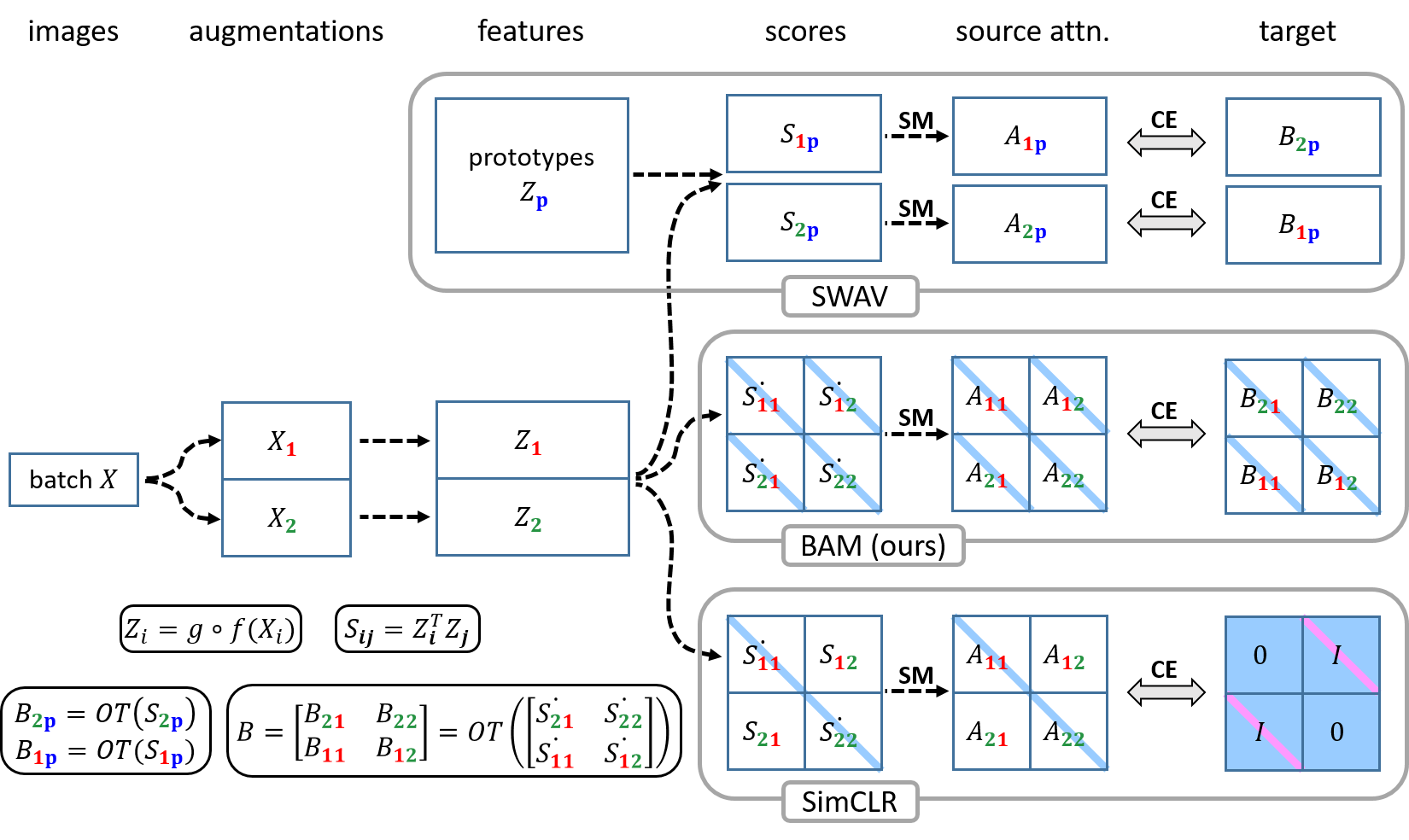}
\vspace{-10pt} 
\caption{
{\small \textbf{Overview of our method, in comparison with \swav \cite{caron2020unsupervised} and \simclr \cite{chen2020simple}}, for the case of $k=2$ views (augmentations). These methods share the principle of converting pairwise features similarities to probabilities, on which a loss is calculated. See text for an overview of the main differences. \green{green} and \red{red} colors refer to first and second views, while \blue{blue} colors refer to prototypes. "SM" refers to the soft-max operation (per row) and "CE" refers to the cross-entropy loss computed between a pair of matrices. Within the matrices, pale blue regions denote zero values while the pink regions represent values of one. The 'dot' above certain blocks of some of the similarity matrices represent zeroing of the diagonal.}}
\label{fig.swav_ours_simclr}  \vspace{-8pt} 
\end{figure*}

Our method can be described in the context of the contrastive \simclr \cite{chen2020simple} and the clustering based \swav \cite{caron2020unsupervised} methods. Refer to Figure~\ref{fig.swav_ours_simclr} for the following explanations. In common to all methods, a batch $B$ of images undergoes $k$ augmentations ($k=2$ in illustration), before undergoing the embedding (encoding and projection) into the latent (feature) space.

Both \algo and \simclr (i) convert the augmented features matrix $Z$ to a pairwise cosine similarity matrix between all augmented images, (ii) apply soft-max to obtain the row-normalized 'attention' matrix $A$, and (iii) compare $A$ to a 'target' matrix using cross-entropy loss.
Critical differences are: 

\vspace{-7pt}
\paragraph{- Target is balanced and entropy-regularized (\algo) vs. binary 'pos'/'neg' (\simclr).} 
The \simclr loss implements the contrastive approach of treating augmented pairs as 'positives' and all other pairs of "negatives", hence the target matrix is taken to be the all zeros matrix, with ones at the positive pair locations (not including self-similarities on the main diagonal). As was claimed and shown empirically in the paper, the "negative" pairs contain interesting relative information (see Figure 1 in main paper) that can be exploited by metric learning methods. \algo constructs a target matrix by taking a balanced and entropy-regularized version of the attention matrix, such that each source matrix row is compared to the regularized matrix row that corresponds to an augmented version of the same image (which can be seen in the illustration by the swapping of the augmentation indices between source and target). This richer kind of supervision drives the features to more complex interactions, based on our main observation that augmented images should have similar self-attention distributions. Furthermore, the control on the entropy (through the gap that we maintain between entropies of source and target matrices) drives the embedding to produce attention distributions that are more 'decisive' (i.e. with lower entropy).

\vspace{-7pt}
\paragraph{- Loss on 'negatives' (\algo) vs. on 'positives' (\simclr).} Since \simclr treats augmented pairs as 'positives' and others as negatives, its loss takes into account only the $2n$ positive pair attention values (which are compared to the pink ones on the diagonals). \algo, on the other hand, does the opposite. As was presented in the paper (Sec. 3.1), the self-attention matching can be done over the entire set of pairs (excluding only the main diagonal self-pairs). However, as discussed in Sec. 3.2 ("Focus on negatives"), we avoid working with distributions that are dominated by the augmented pairs, by zeroing the positive pair similarities (step 14 in Alg. 1), which leads to a general up-weighting of the "negative" pair attention values, due to the row-only (soft-max) and row-column (Sinkhorn) normalizations. This change ("pos"-masking) was ablated in Table 1 of the paper, showing its high significance. Note that even after this masking, the $\Theta(n^2k^2)$ "negatives" out-number $\Theta(nk^2)$ "positives" for typical settings (where $n>1000$ while $k<10$), providing a better supervision signal.

\algo bares many similarities with the clustering based \swav, that introduced a swapped prediction mechanism, which our loss in inspired by.
In addition, like most current instance discrimination self-supervised learners, we build on the multi-crop technique of by \swav. The major differences are:

\vspace{-7pt}
\paragraph{- Attention is batch-to-itself (\algo) vs. batch-to-prototype (\swav).} 
The \swav method, like several other self-supervised representation learning methods (e.g. \cite{mitrovic2020representation}) require maintaining and updating data across the scope of a batch. \swav learns, in conjuction with the embedding, a rough clustering of the training set, in the form of cluster centers, which are updated by each batch through the loss function. In comparison with our method that operates entirely on the pairwise distances between augmented batch images, the \swav approach incurs additional computation and memory, and must make some assumptions on the number of classes in the dataset. In fact, it uses a number of clusters that is several times the actual number of classes and is limited to working with large batch sizes (compared to the number of clusters) in order to obtain meaningful normalized attention maps between the batch and the clusters.

\vspace{-7pt}
\paragraph{- Normalization is 'global' (\algo) vs. 'local' (\swav).}
Similarly to \swav, we compute cross entropy losses between soft-max based attention and optimal-transport regularized attention. However, \algo is unique in that the optimal-transport regularization (obtained by the Sinkhorn algorithm) is computed \textit{once}, in a global manner over the entire very large $kn\times kn$ matrix, as opposed to such normalizations, that are computed $k$ times separately ($k=2$ in illustration of Fig.~\ref{fig.swav_ours_simclr}), once per augmentation. As was empirically shown in Table 1 of the paper, this "global normalization" setting brings an important contribution to our results.




\section{A Simple Clustering Experiment}\label{sec:Cluster_exp}
In order to show the advantage of our embeddings in a direct manner that does not involve training a supervised classifier on top of the feature embeddings, we ran a large-scale unsupervised clustering experiment, where a standard k-means procedure is applied to the embeddings of \algo in comparison with those of \simclr and \swav. For fairness in comparison, all three methods were trained on the ImageNet train set with batch-size of 4096 for 200 epochs. 

After training the embeddings, we extract the respective features of the entire ImageNet validation set. Then, we create clustering tasks, by choosing $t$ classes at random (out of the 1000 classes) and collecting the entire set of images of these classes (with an average of 50 images per class). We created 10 such random clustering instances, for each value of $t$ (number of classes) in the set $\{10, 20, 40, 80, 160\}$.

\begin{figure}[t!]
\hspace{5pt}
\vspace{-10pt}
\includegraphics[width=0.88\columnwidth]
{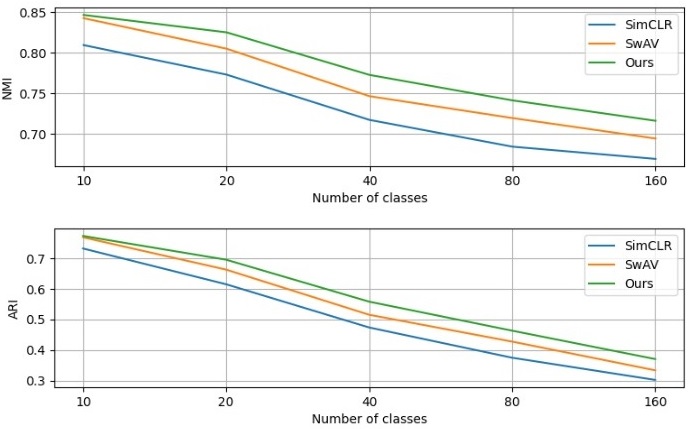}
\vspace{4pt} 
\caption{
{\small \selectfont
{\textbf{Clustering on ImageNet~\cite{ILSVRC15} validation set using k-means over the different feature embeddings.} }}}
\label{fig.clustering}  \vspace{-2pt} 
\end{figure}

The averaged clustering measures of NMI (Normalized-Mutual-Information) and ARI (Adjusted-Rand-Index) are plotted in Figure~\ref{fig.clustering}, showing a consistent inherent advantage of the \algo embedding, across the different instance sizes.

\end{document}